\documentclass[conference]{IEEEtran}

\usepackage{graphicx}
\usepackage{hyperref}
\usepackage{listings}
\usepackage{booktabs}
\usepackage{multirow}
\usepackage{amsmath}
\usepackage[most]{tcolorbox}
\usepackage{xcolor}
\usepackage{booktabs} 
\usepackage{amsmath}
\usepackage{amsthm}
\usepackage{multirow,tabularx,ragged2e}
\usepackage{adjustbox}
\usepackage{graphicx}
\usepackage{enumitem}
\usepackage{lipsum}
\usepackage{multirow}
\usepackage[caption=false, font=footnotesize]{subfig}

\usepackage{booktabs}
\usepackage{makecell}   
\usepackage{adjustbox}
\usepackage[most]{tcolorbox}
\usepackage{cuted}
\usepackage{orcidlink}

\usepackage{hyperref}

\begin{document}
%
\title{Synthesizing Human Trajectory Anomalies: An LLM-Driven Framework with Map-Constrained Routing and Sensor Noise Simulation}

\author{
    \IEEEauthorblockN{Yueyang Liu}
    \IEEEauthorblockA{\textit{Department of Computer Science} \\
    \textit{Emory University}\\
    Atlanta, USA \\
    yueyang.liu@emory.edu}
    \and
    \IEEEauthorblockN{Joon-Seok Kim}
    \IEEEauthorblockA{\textit{Department of Computer Science} \\
    \textit{Emory University}\\
    Atlanta, USA \\
    joonseok.kim@emory.edu}
    \and
    \IEEEauthorblockN{Andreas Z{\"u}fle}
    \IEEEauthorblockA{\textit{Department of Computer Science} \\
    \textit{Emory University}\\
    Atlanta, USA \\
    azufle@emory.edu}
}

\markboth{Journal of \LaTeX\ Class Files,~Vol.~14, No.~8, August~2015}%
{Shell \MakeLowercase{\textit{et al.}}: Bare Demo of IEEEtran.cls for IEEE Journals}
%



\maketitle

\begin{abstract}
Although the study of human trajectory anomalies is vital for spatial data mining, empirical research remains constrained by the lack of ground-truth datasets.
Despite the availability of several real-world and simulated human trajectory datasets, most only include normal mobility patterns and lack annotated anomalies.
This limitation primarily arises because anomalous events are rare in the real world and difficult to collect.
Furthermore, collecting large-scale mobility data is hindered by strict privacy regulations and acquisition costs.
To overcome these limitations and establish an explainable human trajectory anomalies dataset with annotated ground truth, we introduce an end-to-end generative framework designed to synthesize context-aware trajectory anomalies at scale.
Unlike existing models, our framework applies real-world physical constraints on baseline simulated trajectories.
We employ Large Language Model (LLM) agents to systematically inject semantically meaningful behavioral anomalies such as irregular out-of-distribution check-ins and skipped routine visits.
To ensure the spatial validity, the system leverages a map-constrained routing module to recalculate the physical transitions between these LLM agent-modified staypoints.
Moreover, to narrow the simulation-to-reality gap, we augment the resulting trajectories with a context-aware spatial noise model, parameterized by environmental and location-specific variables, to emulate heterogeneous GPS sensor degradation.
Through this pipeline, our framework produces a CTI \& EBI formatted mobility dataset with annotated anomalies, providing a benchmark for downstream spatial data mining tasks.
\end{abstract}

\begin{IEEEkeywords}
Trajectory Data, Anomaly Generation, Human Mobility, GPS Noise Simulation, Large Language Models
\end{IEEEkeywords}

%
\IEEEpeerreviewmaketitle

\section{Introduction}
\label{sec:introduction}
Human mobility data is used in many applications such as trajectory prediction~\cite{10.1145/3748636.3771319}, point of interest (POI) popularity estimation~\cite{chang2021mobility}, and pandemic modeling~\cite{gao2020mapping,pesavento2020data}.
While prior work focuses on normal movement, emerging research shows growing interest in detecting when individuals deviate from established routines ~\cite{10.1145/3681765.3698463,hosseini_sereshgi_semantic_2025}. Examples of such deviations include (1) a child detected to miss school, which may trigger a notification to parents, or (2) a group of commuters from an area staying at home, which may be an early indication of an infectious disease outbreak. Such deviations can be automatically detected from GPS data as trajectory anomalies which have been studied thoroughly in the ICDE community~\cite{li2024causaltad,liu2020online,zhang2023online,lee2008trajectory} and in other literature~\cite{stanford2024numosimsyntheticmobilitydataset,amiri2024urban}.  
Despite the practical applications, the field of trajectory anomaly detection is limited by a lack of anomaly-labeled datasets, which are difficult to acquire for two main reasons. First, because anomalous events are very rare and difficult to manually annotate. Second, location privacy requirements conceal sensitive movements, leaving many anomalous deviations absent or unlabeled in real-world data.

\begin{figure*}[!t]
    \centering
    \subfloat{%
        \includegraphics[width=0.5\textwidth,trim=0cm 0cm 0cm 1.21cm,clip]{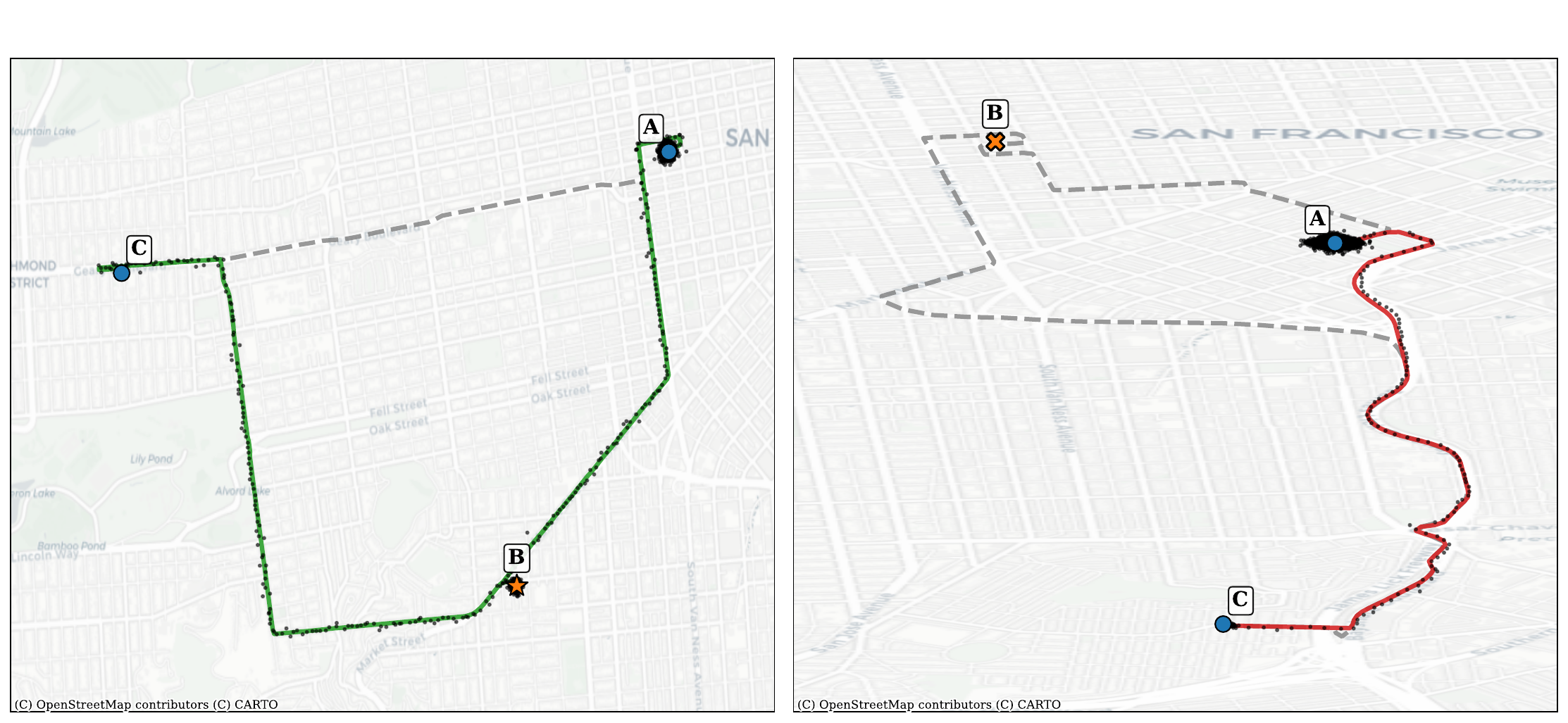}%
    }%
    \subfloat{%
        \includegraphics[width=0.5\textwidth,trim=0cm 0cm 0cm 1.2cm,clip]{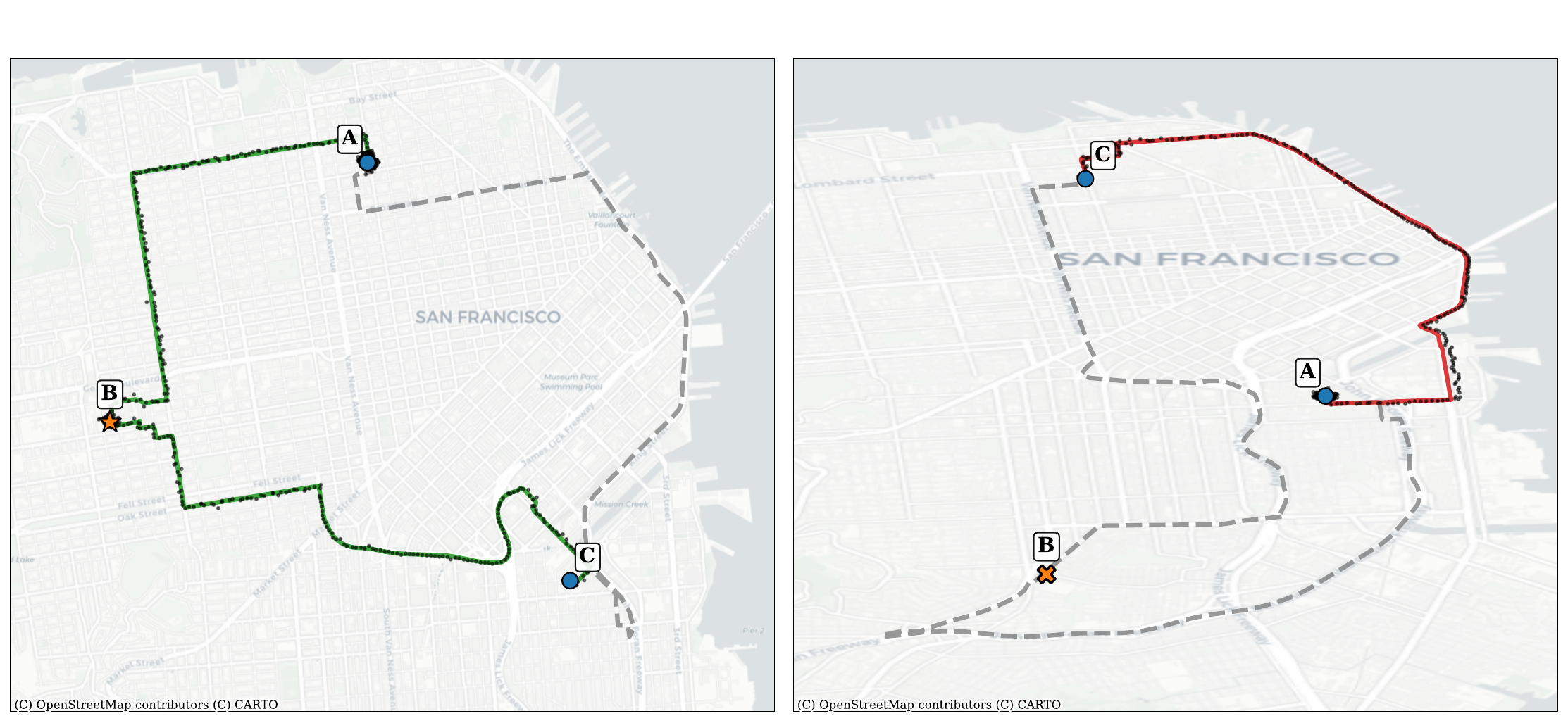}%
    }%
    \vspace{-0.2cm}
    \caption{Examples of synthetic trajectory modification: Insertion of new trips to a new Location~B in-between two Locations~A~\&~C (1st and 3rd subfigure) and removal of visited Location B creating a direct trajectory between Locations~A~\&~C (2nd and 4th subfigure). Dashed grey lines denote the original trajectory. Solid green and red lines denote trajectories generated as described in this work. Black dots along trajectories denote generated data after adding noise.}
    \label{fig:samples}\vspace{-0.5cm}
\end{figure*}
Existing public mobility datasets do not solve this problem of a lack of annotated anomalies for trajectory data. Current large-scale human mobility datasets range from semantic point of interest (POI) check-ins (e.g., Gowalla~\cite{10.1145/2020408.2020579}, Foursquare~\cite{6844862}, Massive-STEPS~\cite{wongso2026massivestepsmassivesemantictrajectories}) to coordinate traces (e.g., GeoLife~\cite{zheng2011geolife}, T-Drive~\cite{zheng2011t-drive}, Grab-Posisi~\cite{10.1145/3356995.3364536}, WorldTrace~\cite{unitraj2025}). While these datasets provide individual-level trajectories at various spatial and temporal granularities, none of these datasets captures ground-truth anomaly labels.
Still, such trajectory datasets without any anomaly annotation or ground truth are commonly used for the task of trajectory anomaly detection. The seminal work of Lee, Han, and Li~\cite{lee2008trajectory} does not use any ground truth, and validates anomalous results only qualitatively. 
More recently, authors have augmented trajectories with synthetically generated anomalies noting that ``Since no labeled dataset is available for trajectory anomaly detection [...] we generate anomalous trajectories for evaluation''~\cite{li2024causaltad} and ``Since there is no labeled dataset available for anomalous trajectory detection [...] we use generated anomalies for the evaluation''~\cite{liu2020online}. Other related works use a cohort of human annotators to identify anomalous routes~\cite{zhang2023online} which is not scalable beyond more than a few thousands of labeled routes (1,688 in ~\cite{zhang2023online}). The existing works, all published at ICDE or TKDE, suffer from the same limitation: A lack of available benchmark data of trajectory data with anomaly labels. The goal of this work is to address this gap.

Recently, synthetic datasets with anomaly labels have been generated through agent-based simulation. The NUMOSIM~\cite{stanford2024numosimsyntheticmobilitydataset} is explicitly designed to provide native labels for benchmarking abnormal behaviors. 
NUMOSIM’s anomalies are confined to random alterations such as shifting a stay point's duration or inserting an unseen location without causal reasoning as to why the anomaly occurred.
In addition, NUMOSIM does not provide raw continuous trajectories alongside its discrete check-ins, leaving a critical void in the benchmarking of continuous spatial anomaly analysis systems. Another dataset of anomalies trajectories is Urban Anomalies~\cite{amiri2024urban} in which the anomalies are caused by changes in the behavior of agents, such as skipping work, or eating more frequently. Like NUMOSIM, this work only provides anomalies at the semantic level, but the movement trajectories between places (normal or anomalous) follow a shortest path without anomalies. 

To overcome these limitations, this paper introduces a cost-effective generative framework that leverages LLMs to synthesize targeted, context-aware trajectory anomalies.
Recognizing the sequential structural similarity between natural language and human mobility, our methodology employs a persona-driven LLM acting as a ``High-level Reasoning Agent" to profile each agent’s routine and plan anomalies that break this established routine, rather than inserting anomalies at random as in NUMOSIM, while keeping them physically feasible via Insert and Skip strategies, as illustrated in Figure ~\ref{fig:samples}. 

Furthermore, to overcome the challenge of LLMs hallucinating spatial coordinates, we introduce a robust three-stage translation architecture. 
By integrating OSM-based POI resolution, demographics-aware travel-mode prediction, and map-constrained routing, this framework sequentially converts semantic anomaly sequences into continuous, 0.2Hz raw GPS coordinate traces. 
Finally, to effectively close the simulation-to-reality gap, we apply a multi-layered noise model to emulate complex real-world sensor imperfections.

Specifically, this paper makes the following contributions:
\begin{enumerate}[leftmargin=*]
    \item Novel LLM-Driven Framework for Anomaly Synthesis: We introduce a persona-driven anomaly generation framework that utilizes LLMs to analyze human mobility routines and generate contextual, persona-consistent anomalies. The system employs targeted Insert and Skip strategies while strictly enforcing physical feasibility parameters to prevent spatial-temporal overlaps.
    \item Robust Translation from Semantic Sequences to Continuous Raw Coordinates: To mitigate LLM spatial hallucination, we developed a three-stage conversion architecture. This includes an OSM-based POI resolution step, a rule-based travel-mode predictor based on demographics and trip distance, and precision routing via Valhalla to synthesize continuous, 0.2Hz GPS coordinate traces.
    \item Multi-Layered Noise Simulation for Sim-to-Real Transfer: To narrow the gap between Valhalla generated road centerlines and empirical GPS observations, we design a multi-layered noise injection framework that simulates macro-scale drift, building-induced multipath reflections, and signal dropouts, ensuring downstream detectors are evaluated against realistically coordinates rather than geometrically sanitized paths.
\end{enumerate}
Our generated dataset based on SF-Life, perturbed by anomalies and noise, is available at  \url{https://huggingface.co/datasets/spatialcomputing/TPAD}
\vspace{-0.2cm}
\section{Background and Related Work}
\vspace{-0.0cm}
\label{sec:related_works}
In this section, we review the existing literature and foundational concepts that inform our generative framework. 
\vspace{-0.2cm}
\subsection{Human Trajectory Anomalies} \label{sec:related_paper}
While an anomaly is generally defined as a data point that significantly deviates from a normative data distribution, the specific definitions of human trajectory anomalies vary considerably across studies. Consequently, rather than establishing a singular foundational definition, this section surveys recent geospatial literature to systematically categorize the diverse types of human mobility anomalies. Recent literature generally categorizes trajectory and mobility anomalies into five distinct paradigms based on spatial, temporal, semantic, and relational contexts. 
\begin{enumerate}[leftmargin=*]
\item \textbf{Sub-trajectory and Route Anomalies:} Anomalies frequently manifest as localized spatial deviations within a specific trip, rather than encompassing the entire trajectory. This classification is primarily composed of two structural types: detour anomalies and switch anomalies~\cite{jiang_recast_2025}. Detour anomalies occur when an agent navigates an unnecessarily elongated or alternative route, typically to avoid sudden roadblocks or unexpected traffic. Switch anomalies occur when multiple valid routes exist between a given source-destination pair and a trajectory initially follows one established normal route but abruptly switches to a completely different normal route in the middle of the trip. Additional anomalies include Time Anomalies, in which the travel speed of a trip changes and Loop Anomalies in which loops from a location to itself are added. 
\item \textbf{Individual Behavioral and Routine Anomalies:} Aligning with the most commonly understood definition of behavioral irregularities, these anomalies represent temporal and spatial deviations from an individual agent's established historical ``patterns of life''~\cite{hosseini_sereshgi_semantic_2025,amiri2024urban}. Within this classification, an abnormality is strictly relative to the specific individual rather than a generalized population baseline. It fundamentally defines an anomaly as an instance where an agent visits a location they typically do not frequent or abruptly breaks a recurring schedule. 
\item \textbf{Contextual and Semantic Anomalies:} These anomalies are defined as structurally valid movements that only become anomalous when evaluated at the intersection of an agent's specific identity, temporal variables, and the semantic category of the visited POI~\cite{duan_back_2024}. The anomalous state of a trajectory is strictly conditioned upon the specific demographic or occupational role of the agent. 
\item \textbf{Collective and Co-traveling Anomalies~\cite{wen_cobad_2025}:} This classification shifts the analytical focus from isolated agents to complex spatial-temporal dependencies and interactions among multiple individuals. A trajectory may appear normative individually but become highly anomalous when contextualized against the expected co-occurrence of social ties. Literature categorizes these interaction-driven deviations into three primary subtypes: unexpected co-occurrence anomalies, where agents share a location without historical precedent; absence anomalies, where an expected participant is missing from a joint event; and synthetic coordination anomalies, where historically isolated individuals artificially converge at a shared destination. 
\item \textbf{Kinematic and Physical Anomalies~\cite{sharma_towards_2025,kim2025grounded}:} This classification defines an anomaly as a spatial-temporal sequence that violates fundamental physical laws or mechanical constraints, such as implausible acceleration or teleportation. In the context of human mobility, pure physical anomalies are exceedingly rare. Consequently, these deviations are primarily treated as issues of underlying data quality rather than reflections of actual human behavior. 
\end{enumerate}
\vspace{-0.0cm}
\subsection{Human Mobility Anomaly Datasets}\label{sec:related_db}
In this section, we examine several prominent human mobility datasets.
Within the spatial data mining literature, human mobility datasets predominantly rely upon four distinct data representation frameworks: constant time intervals (CTI), variable time intervals (VTI), event-based intervals (EBI), and constant time periods (CTP)~\cite{kong2024human}, with constant time intervals and event-based intervals being the most widely adopted.

\subsubsection{Constant Time Intervals (CTI) representations} This representation methods consist of spatial coordinates, specifically latitude and longitude, paired with exact timestamps at constant intervals. These trajectories are typically captured through GPS-enabled devices and theoretically represent a continuous and smooth movement path. This continuous format explicitly captures the moving object's kinematic state at any given moment, including velocity, acceleration, and rate of turn.  

YJMob100K~\cite{yabe_yjmob100k_2024} provides an extensive, longitudinal record of human mobility CTI formatted trajectories, capturing the movements of up to 100,000 individuals over a 75-day period within four metropolitan areas. The raw location pings are severely discretized into spatial grid cells and 30-minute temporal bins. While the dataset includes a 15-day subset captured during an undisclosed emergency, it entirely lacks the explicit, ground-truth anomaly labels required for training and benchmarking supervised trajectory anomaly detection models.

Urban Anomalies~\cite{amiri2024urban} is generated via an agent-based simulation. It provides data in both continuous raw coordinate formats and discrete semantic representations, encompassing staypoints and check-ins. 
The framework systematically injects predefined behavioral deviations across four distinct categories: increased feeding frequency, altered recreational preferences, randomized recreational site selection, and occupational absenteeism. 
Despite native ground-truth annotations, the framework assumes constant-speed, shortest-path routing that ignores empirical kinematic variances. Additionally, its outliers are restricted to basic behavioral scheduling rather than spatial routing anomalies, allowing classic algorithms to detect these structural deviations~\cite{10.1145/3681765.3698463}.

\subsubsection{Event-Based Intervals (EBI) Representations}
Event-Based Intervals representations provide a higher-level abstraction of mobility, where a continuous trajectory is discretized into a chronological sequence of stay points. Rather than passively logging raw physical coordinates, this format effectively functions as a targeted check-in table restricted to a limited set of POI locations. These discrete points are usually enriched with contextual and semantic metadata, explicitly identifying the functional purpose of the location alongside precise arrival and departure timestamps. 
Currently, NUMOSIM~\cite{stanford2024numosimsyntheticmobilitydataset} stands as the one of a few EBI formatted trajectory datasets equipped with explicit ground-truth annotations for anomaly detection. It comprises synthetic mobility records for 200,000 agents over an 8-week period in Los Angeles. 
Specifically, it introduces two primary categories of anomalies: altered stay points: characterized by temporal shifts in arrival or departure times, and inserted stay points: representing novel spatial visits not observed during the training phase. 
\vspace{-0.2cm}
\subsection{LLM-based Human Trajectory Generation}\label{sec:related_llm}
Recently, the rapid advancement of LLMs has opened new avenues for human mobility generation, overcoming the semantic and interpretability limitations of traditional deep generative models. 
In the Geo-Llama~\cite{li2025geo} framework, the LLM is fine-tuned for trajectory generation by formulating the task as a next-token prediction problem, encoding input trajectories as standardized text strings. 
Conversely, MobGLM~\cite{zhang2024mobglm} uses multimodal tokens that represent demographic attributes, location zones, activity types, and transportation modes. CAMS~\cite{du_cams_2025} generates trajectories by utilizing user demographic profiles, regional summaries, and historical activity sequences as inputs. 
TrajLLM~\cite{ju2025trajllm} iteratively predicts the next logical semantic activity using demographic data, personality traits, POIs, and historical routines.
Finally, LLMob~\cite{wang2024large} incorporates personas, specific contextual prompts, and historical data, utilizing retrieval-augmented reasoning to infer daily motivations and combine them with identified habitual activity patterns to generate daily activity trajectories.  

A primary challenge in LLM-driven spatial generation is validating outputs and mitigating spatial hallucinations (e.g., physically implausible or topologically incorrect).
While various mechanisms exist to address this, they differ across models.
For example, Geo-Llama~\cite{li2025geo} relies on a post-processing integrity check to resolve physical impossibilities. CAMS~\cite{du_cams_2025} addresses geographic hallucinations by tracking the Toponym Valid Ratio metric, utilizing fine-grained POI data, and employing hierarchical address representations. MobGLM~\cite{zhang2024mobglm} avoids generating fake coordinates by confining the spatial generation to predefined administrative small zone codes and measures sequence validity using automated metrics. 
TrajLLM~\cite{ju2025trajllm} prevents spatial inaccuracies by strictly limiting LLM-based recommendations using real POI data and historical visit frequencies. LLMob~\cite{wang2024large} scores the generated patterns by comparing them against real-world trajectories to ensure consistency.

Despite these mitigation strategies, LLM hallucination remains a significant challenge in ensuring spatial validity. Consequently, the majority of current models predominantly generate semantic or check-in-level records rather than fine-grained, raw trajectory data. Furthermore, the advancement of this domain is constrained by accessibility, as several of the datasets utilized in existing literature remain closed-source.

\vspace{-0.2cm}
\section{Methodology}\vspace{-0.1cm}
\label{sec:methodology}
\begin{figure*}[t]
    \centering
    \includegraphics[width=.99\linewidth,trim=0cm 8.1cm 2.2cm 0.0cm,clip]{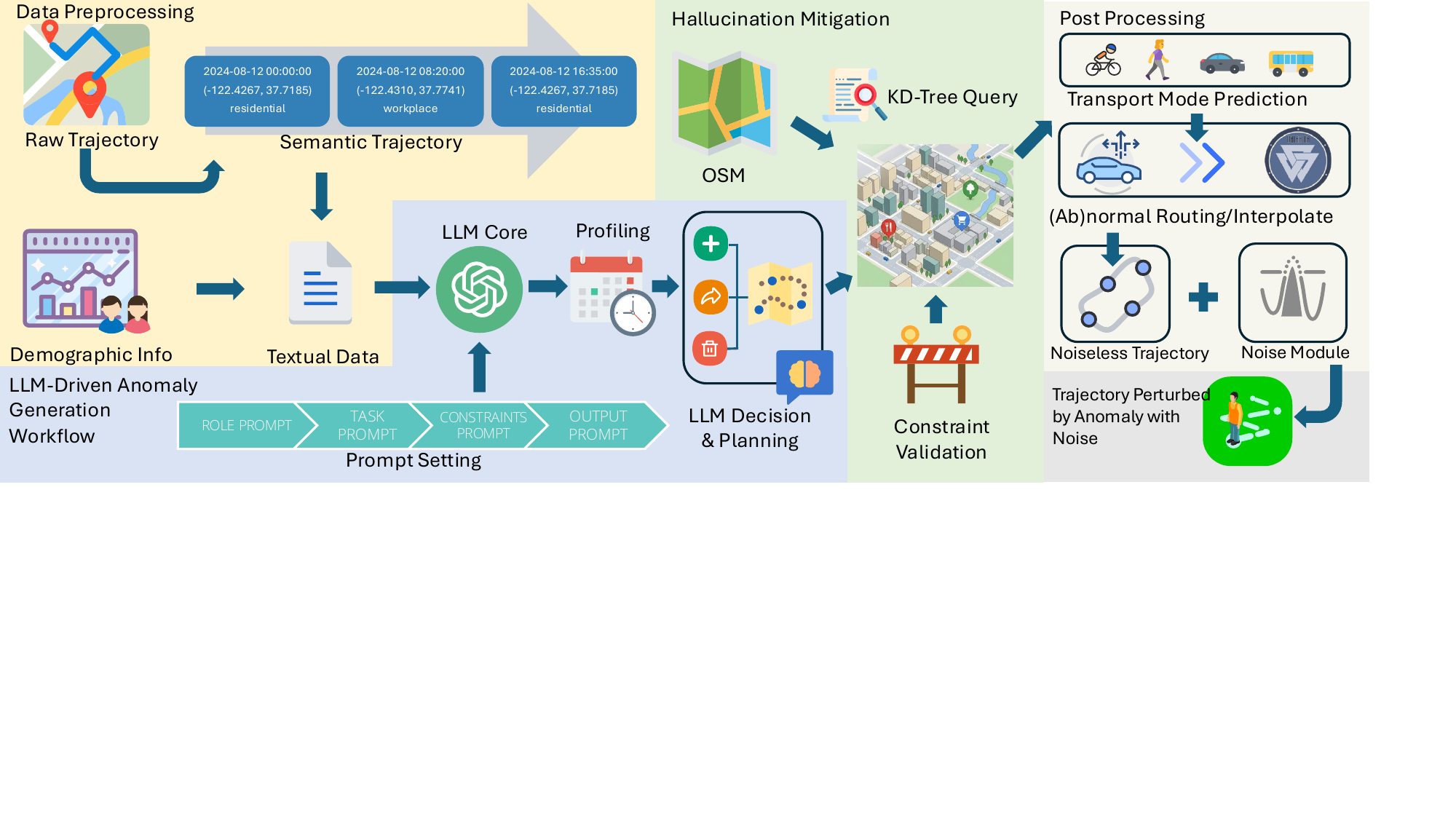}
    \vspace{-0.0cm}
    \captionsetup{width=1.\linewidth}
    \caption{The Illustration of Our Framework for Synthesizing Anomalous Trajectories. (Left) Data Preprocessing; (Middle Lower) LLM-Driven Anomaly Generation; (Middle Upper) Hallucination Mitigation; (Right) Post Processing.\vspace{-0.45cm}}
    \label{fig:framework}
    \vspace{-0.2cm}
\end{figure*}
To address the scarcity of labeled anomaly data in spatial-temporal domains, we propose a novel end-to-end framework for synthesizing anomalous trajectories that produces dual-format datasets with both CTI and EBI.
In this section, we detail the deployment of an LLM as the spatial-temporal anomaly generator. 
Specifically, the generation pipeline is partitioned as follows: the Valhalla routing engine~\cite{githubGitHubValhallavalhalla} routes through injected POIs, producing route-level detour anomalies (a subset of Sub-trajectory and Route Anomalies), whereas the LLM directly formulates the Individual Behavioral and Routine Anomalies, as well as the Contextual and Semantic Anomalies. 
In addition, to mitigate LLM spatial hallucinations and overcome the sparsity of traditional check-in records, our approach implements a dual-stage generation mechanism: OSM-based POI resolution and continuous, 0.2 Hz (5-second) trajectory generation. 
The whole framework is shown in Figure~\ref{fig:framework}, which includes four components: Data Preprocessing, LLM-Driven Anomaly Generation, Hallucination Mitigation section, and Post Processing for raw trajectory synthesis.
Furthermore, unlike recent work, we introduce a training-free, environment-aware noise model to enhance the realism of the synthesized trajectories.

\vspace{-0.3cm}
\subsection{Data Ingestion and Preprocessing}\label{processing}
\vspace{-0.1cm}
The data collection and preprocessing pipeline transforms raw tabular mobility data into a structured textual format optimized for LLM processing.
In this paper, we primarily utilize the SF-Life~\cite{algama2026sflifelargescalesimulatedmovement} dataset check-in record part and Foursquare NYC~\cite{6844862} data. 
To minimize memory overhead and maximize computational efficiency, the preprocessing module selectively extracts the five essential features required for spatial-temporal modeling: User ID, Timestamp, Latitude, Longitude, and POI Category. 
Let $U = \{u_1, u_2, \dots, u_{|U|}\}$ denote the set of users. For a specific user $u \in U$, a raw spatial-temporal trajectory is defined as a chronologically ordered sequence of check-in events, denoted as: $T_u = \langle e_1, e_2, \dots, e_n \rangle$.
Each event $e_i$ is represented as a tuple $e_i = (t_i, c_i, l_i)$, where: $t_i$ is the chronological timestamp. $c_i$ is the categorical Point of Interest (POI) string. $l_i = (\phi_i, \lambda_i)$ represents the spatial coordinates, consisting of latitude $\phi_i$ and longitude $\lambda_i$.
Because LLMs require textual inputs, the tabular trajectory data is subsequently mapped into a continuous semantic representation. 
The dataset is grouped by User ID, and each individual check-in event is serialized into a strict, standardized formatted textual string: $s_i = t_i | c_i | @ \phi_i, \lambda_i$, where '$|$' is a delimiter.
The final semantic trajectory for user $u$, denoted as $\mathcal{S}_u$, is constructed by concatenating the ordered sequence of strings $\{s_1, s_2, \dots, s_n\}$.
Furthermore, enabled by the detailed metadata provided in the SF-Life dataset, we integrate agent demographic profiles into the modeling pipeline. This demographic context includes the agent's age, gender and home location.

\subsection{LLM-Driven Anomaly Generation}

Our framework employs an LLM to generate Individual Behavioral and Routine Anomalies, along with Contextual and Semantic Anomalies. 
To execute these tasks, we develop an agentic workflow that summarizes the agent's historical trajectories and subsequently plans the spatial location and abnormal behavioral patterns of the anomalies. 
Due to computational efficiency and cost optimization concerns, our framework excludes locally deployed architectures and domain-specific spatial models.
Instead, we conducted several experiments across multiple Open AI GPT models to serve as the reasoning backbone for the trajectory generation process.

\subsubsection{LLM Workflow}
Given that GPT model operates as a general-purpose language model rather than a specialized geospatial model, we first design some prompts to steer the LLM toward generating spatiotemporal anomalies. 
To maximize the accuracy and reliability of anomaly outputs, we propose a structured prompt~\cite{heston2023prompt} that includes positive samples and detailed task explanations, ensuring the model correctly parses complex trajectory sequences and fully understands the definition of an anomaly.
Specifically, the prompt architecture can be partitioned into four parts based on their functional roles: role assignment, task definition, critical constraints, and output schema.
Initially, in the role assignment part, we define the operational bounds by assigning the LLM a specialized persona restricted to a predefined spatial and temporal domain. Within this part, we also define the two anomaly strategies that the LLM generates: Insert and Skip.
During the task definition part, we require the model to profile the daily routine of the agent prior to generating a minimum of five distinct anomalies.
Specifically, the LLM first infers the agent type (e.g., a corporate commuter or a university student) by evaluating the daily trajectory sequence alongside the provided demographic data.
Because anomalies are customized to the agent, they are conditioned on past behavioral patterns and designed to intersect with local events, we impose a profiling requirement to establish a routine summary in profile. 
This summary ensures that the downstream generation step produces real behavioral shifts rather than random spatial noise.

Following this profiling step, we require the model to plan at least five distinct anomalies per agent based on their profile. This minimum keeps anomalies rare relative to normal events over the 10-week observation period while remaining within the LLM’s context window limits. These spatial-temporal perturbations are executed by explicitly instructing the model to combine two primary disruption operations: Insert and Skip.
A benefit of this design is that detour anomalies are produced directly by the Insert operation, requiring no dedicated generative architecture.
When a target POI is inserted, it acts as an intermediate waypoint for the Valhalla routing engine, which automatically reroutes the trip through it along the road network.
The operational definitions of these strategies are detailed in Table~\ref{tab:modification_strategies}, and two representative samples are illustrated in Figure~\ref{fig:samples}. 
LLMs were excluded from generating detour path directly due to context window size limitations and a high susceptibility to spatial hallucinations (as described in Section \ref{sec:Hallucinations}). 
Furthermore, because perturbation strategies alter required travel distances, transit times, and transportation modes, 
our framework employs the Valhalla routing engine to compute physically feasible waypoints.
To prevent topologically or temporally implausible trajectories, explicit constraints are specified in the prompt. 
The model must calculate insert windows strictly within existing temporal gaps, ensuring sufficient visit durations without causing event overlaps. 
Finally, to keep the temporal distribution realistic, we specify that a single anomaly may cross midnight into the next day, but an agent's anomalies may not all fall on the same calendar day.
\begin{table}[t]
    \centering
    \caption{Comparison of Trajectory Modification Strategies\vspace{-0.3cm}}
    \label{tab:modification_strategies}
    \renewcommand{\arraystretch}{0.9}
    \begin{tabular}{@{} p{0.08\columnwidth} p{0.82\columnwidth} @{}}
        \toprule
        \textbf{Strategy} & \textbf{Operational Characteristics} \\
        \midrule
        \textbf{Insert} & 
        \textbf{Mechanism:} Synthesizes and injects an out-of-context check-in event into the timeline. \newline
        \textbf{Spatial Target:} Introduces an entirely novel POI or recontextualizes a previously visited coordinate. \newline
        \textbf{Temporal Constraint:} Must fit within an existing temporal gap; requires strict travel-time validation to prevent chronological overlaps.\\
        \midrule
        \textbf{Skip} & 
        \textbf{Mechanism:} Deletes a highly predictable, habitual check-in event from the historical sequence. \newline
        \textbf{Spatial Target:} Removes an expected spatial coordinate without introducing new geographic data. \newline
        \textbf{Temporal Constraint:} Expands an existing temporal gap or transit time; avoids overlapping conflicts.\\
        \bottomrule\vspace{-1.cm}
    \end{tabular}
    \vspace{-1cm}
\end{table}


\subsubsection{Mitigation of Spatial Hallucinations}\label{sec:Hallucinations}

A notable limitation of utilizing LLMs for geospatial tasks is their vulnerability to spatial hallucinations. 
This phenomenon occurs when the model generates highly plausible but entirely fictitious locations, non-existent POI, or topologically impossible coordinates~\cite{wang2025mitigating}. 

To resolve this vulnerability and prevent physically impossible transitions, such as teleportation, our framework decouples the semantic intent of the anomaly from its precise geographic instantiation. 
First, instead of having the LLM output arbitrary location names or raw coordinates, the prompt constrains the model to generate deterministic geographic search parameters.
Specifically, the LLM must output a localized geographic search center, a bounding radius, and categorical OpenStreetMap (OSM) tags. 
Next, a POI-resolution module uses a KD-Tree query on the OSM database to retrieve an existing POI, followed by a constraint-validation step to ensure each insertion respects the temporal gap and is physically viable.
By constraining the semantic output of a large language model to a rigorous geospatial database, this mechanism ensures that every resolved anomalous location physically exists and is topologically valid, thereby eliminating the risk of spatial hallucinations.


Second, we combined a Chain-of-Thought (CoT) reasoning framework directly into the generation pipeline to mitigate hallucinations~\cite{barkley2024investigating,li2025mitigating}. 
Instead of executing zero-shot trajectory modifications, the model is required to decompose the anomaly generation task into a sequence of logical deductions. 
Before defining any executable abnormal generation commands, the LLM must first articulate the established behavioral persona of the agent, isolate the specific chronological pattern targeted for disruption, and explicitly calculate the available temporal insert windows between sequential check-ins.

\subsubsection{Bounding LLM Output for Automated Processing}
To guarantee seamless integration into automated processing pipelines and prevent the generation of extraneous conversational text, the framework constrains the LLM to a strict JSON output schema. By embedding concrete output exemplars directly within the system prompt, we strictly bound the model's generative behavior to a predefined structural format. 
This structural enforcement bypasses the parsing failures common to unstructured text, facilitating reliable and deterministic data extraction for downstream evaluation.



\subsection{Raw Trajectory Synthesis}
While LLMs are utilized to generate continuous mobility trajectories, they frequently fail to capture fine-grained spatial-temporal details~\cite{zhou2025blurred} and introduce movement uncertainty~\cite{han2025bridging}.
To overcome these limitations and ensure the synthesis of strictly valid raw trajectories, the framework employs the Valhalla routing engine. 
Following the derivation of the semantic anomaly parameters, Valhalla is utilized to synthesize the high-resolution, point-to-point coordinate sequences.
To establish the transportation method between sequential POIs, a rule-based predictor first assigns each trip a travel mode and speed based on its spatial-temporal constraints and the agent’s demographics. For example, younger students often favor walking or public transit, provided adequate pedestrian infrastructure and transit availability are present.
Once the optimal travel mode is assigned, the Valhalla engine computes the detailed sub-trajectories and routes the anomalous deviations. 
To guarantee the temporal and spatial accuracy of public transportation routes, we integrate SFMTA~\cite{sfmta_website} General Transit Feed Specification (GTFS) data directly into the routing infrastructure. 
Furthermore, the framework configures the routing engine based on the agent's demographic profile. By mapping these demographic attributes to kinematic constraints and transit preferences, the system adjusts concrete travel metrics such as walking speeds and preferred modes of transport.
This comprehensive integration ensures that the synthesized raw trajectories adhere strictly to real-world transit schedules and the physical capabilities of the simulated agent, thereby preserving the structural realism and kinematic validity of the final mobility dataset.
Crucially, as the Valhalla routing engine translates these LLM perturbations into raw trajectories, it must calculate specific detour waypoints to accommodate the modified spatial dynamics. These waypoints ultimately manifest as the physical sub-trajectory and route-level anomalies.

\subsection{Noise Module}\label{sec:noise}
To resolve the mismatch between idealized road centerlines and real-world GPS observations, we introduce a hierarchical noise generation module. 
This module injects multi-scale coordinate perturbations into nominal trajectories across three distinct phases: macro-scale atmospheric drift, micro-scale urban canyon reflections, and receiver-level signal dropouts.
Without ground-truth baselines for these noise parameters, we calibrate their scaling factors through visual inspection to ensure that spatial deviations remain physically plausible.
\subsubsection{Macro-Environment Noise}
We quantify tropospheric interference using the Saastamoinen delay model~\cite{krueger2005standard,5068392}, which incorporates the zenith delay alongside a simple geometric mapping function. 
Leveraging historical ERA5 weather telemetry~\cite{cds.adbb2d47}, the Zenith Hydrostatic Delay (\textit{ZHD}) and Zenith Wet Delay (\textit{ZWD}) are defined as:$$ZHD=\frac{0.0022768\cdot P}{1-0.00266\cdot\cos(2\phi)-0.00028\cdot H}$$$$ZWD=0.0022768\cdot\left(\frac{1255}{T}+0.05\right)\cdot e_s$$$$SPD=(ZHD\times M_{h}(\epsilon))+(ZWD\times M_{w}(\epsilon))$$
where $P$ is the surface pressure, $\phi$ is the latitude, and $H$ is the altitude. The variable $T$ represents the 2-m air temperature, and $e_s$ denotes the vapor pressure, which is computed using the 2-m dewpoint ($T_d$). To calculate the Slant Path Delay ($SPD$), we apply the hydrostatic mapping function ($M_h(\epsilon)$) and the wet mapping function ($M_w(\epsilon)$). In the absence of explicit satellite ephemeris data, we assume a vertical signal path and set both mapping functions to 1.
Concurrently, signal propagation is subjected to an ionospheric delay ($I$), modeled using the established ionospheric delay algorithm~\cite{klobuchar1987ionospheric}:$$I=\frac{40.3\cdot 10^{16}\cdot TEC}{f_{L1}^2}$$
where $TEC$ represents the Total Electron Content~\cite{GNSS_IGSIONOTEC}, and $f_{L1}$ is the GPS $L1$ signal frequency.
To simulate the continuous temporal evolution of these atmospheric dynamics, the combined delay is formulated as a Gauss-Markov AR(1) process~\cite{gallon2021robust}. Given a data sampling interval $\Delta t$ and a relaxation time constant $\tau=300$ s, the sequential delay state $X_i$ updates recursively:
$$X_i = e^{-\frac{\Delta t}{\tau}} \cdot X_{i-1} + \mathcal{N}\left(0, \left(1 - e^{-\frac{2\Delta t}{\tau}}\right)\left[(SPD_i + I_i)\right]^2\right)$$

\subsubsection{Micro-Environment Noise}
The second layer models the multi-path noise caused by dense surrounding buildings. To model the multi-path scattering effects characteristic of dense urban infrastructure, we inject localized positional noise scaled by a zero-mean Laplace distribution~\cite{liu2024overbounding}:
$f(\epsilon_{mp, i}; b_i) = \frac{1}{2b_i} \exp(-|\epsilon_{mp, i}|/b_i)$.
To dynamically modulate this interference based on spatial topology, we leverage structural footprint geometries extracted from OSM. For any given ground-truth receiver state $X_i$, we compute the minimum Euclidean distance $d_i$ to the nearest building polygon. We establish a 15-meter proximity buffer to quantitatively delineate the urban canyon boundary.
For open environments, the scale parameter $b_i$ is set to 3; when a trajectory intersects the urban canyon buffer, it scales to 15 to emulate severe signal scattering. 
\subsubsection{Receiver System Noise}
The final layer models internal receiver noise and external signal blockage. Thermal interference within the hardware is represented as standard Gaussian noise with a standard deviation of $\sigma=1.5$. 
To simulate severe signal blockages, complete dropouts, or momentary receiver failures caused by physical obstructions, we implement a dynamic masking function governed by a Poisson process. 
Let $M_i$ denote the binary blocking mask at epoch $i$, where the masking probability is dictated by the threshold parameter $\lambda$. 
The probability mass function for the underlying Poisson event occurrences $k$ is given by: $P(k; \lambda) = \frac{\lambda^k e^{-\lambda}}{k!}$ , such that a blockage occurs ($M_i=1$) when $k>0$.
The active signal input mask threshold $\lambda$ is dependent upon the physical environment, specifically open conditions versus highly obstructed topologies, such as tunnels.
During intervals of active signal loss, we model these severe positional outliers, $c_i$, by sampling from a heavy-tailed Cauchy distribution: 
$f(c_i; x_0, \gamma) = \frac{1}{\pi\gamma[1+((c_i-x_0)/\gamma)^2]}$
where the location parameter $x_0$ is centered at $0$, and the scale parameter $\gamma$ is aggressively set to 15 (truncated to $\pm 200m$) to simulate the heavy tails characteristic of complete signal degradation. 

We notice that a recently published work, DeTraj~\cite{10.1145/3849730.3849829}, also injects noise into trajectories. However, DeTraj relies on stay-point detection, although stay-point labels commonly do not exist in real-world data; it also requires training supervised models on large-scale real-world trajectories. Instead, we focus on environment-based noise that requires no stay-point detection or training.

\section{Experimental Results and Analysis}
\label{sec:results}
\vspace{-0.0cm}
To systematically evaluate the spatial-temporal validity, and downstream applicability of our proposed anomaly synthesis framework, we conduct an empirical analysis on both continuous trajectory and check-in datasets. Specifically, our experimental design is structured to address the following four core research questions:
\begin{itemize}[leftmargin=*]
    \item \textbf{Q1 Realism and Plausibility:} Does the proposed framework generate trajectory anomalies that preserve kinematic realism and schedule plausibility, while demonstrating a contextual understanding of agent profiles and logical selection for spatial perturbations? 
    \item \textbf{Q2 LLM Backbone Efficiency:} How does the choice of LLM backbone impact token consumption, cost, and overall generation performance during pipeline execution?
    \item \textbf{Q3 Spatial Hallucination Mitigation:} Can the combination of OSM tags and model-generated search radii eliminate spatial hallucinations compared to direct POI search, and how do varying search radii from different LLMs impact this outcome?
    \item \textbf{Q4 Downstream Benchmark Utility:} Can current anomaly detectors successfully differentiate the generated behavioral anomalies from normal trajectories, and how does detection performance on the generated dataset compare with an existing synthetic benchmark?
\end{itemize}
\subsection{Experimental Setup}
\subsubsection{Datasets}
\label{sec:dataset}
We conduct extensive experiments leveraging two distinct spatial-temporal datasets: SF-Life dataset and Foursquare NYC dataset. 


To ensure a balanced and computationally feasible evaluation, we select a representative sample of 100 unique users from SF-Life (5-second resolution; $\approx 1.2 \times 10^6$ points per agent) and the 100 most active users from Foursquare NYC.

\subsubsection{Data Processing and Prompt Serialization}
To operationalize the raw tabular check-in data for LLM inference, we apply the tabular serialization pipeline, detailed in Section \ref{processing}, which maps discrete spatial-temporal records into structured context windows. To optimize token utilization and maximize the effective context length of the LLM, the continuous spatial coordinates $(\phi, \lambda)$ are quantized to a precision of four decimal places. This lexical quantization preserves sufficient spatial resolution for urban mobility modeling while significantly reducing token cost. 


\begin{table}[t]
\centering
\caption{Quantitative Overview of Anomaly Perturbation \& Spatial Mapping 
}
\vspace{-0.2cm}
\label{tab:LLM_Perturbation_statistics}
\resizebox{\columnwidth}{!}{%
\begin{tabular}{@{}llcccc@{}}
\toprule
\makecell[l]{\textbf{Dataset} \\ \textbf{(OSM Search}  \textbf{Strategy)}} &
\textbf{Model} &
\makecell[c]{\textbf{Inserts} \\ \textbf{Mapped/} \\ \textbf{Planned}} &
\makecell[c]{\textbf{Planned} \\ \textbf{Skips}} &
\makecell[c]{\textbf{Insert} \\ \textbf{Resolve} \\ \textbf{Rate}} &
\makecell[c]{\textbf{Median} \\ \textbf{Matched} \\ \textbf{Radius (m)}} \\
\midrule
\multirow{3}{*}{\makecell[l]{SF-Life  (Tag)}}
 & GPT-5-mini   & 337/368 & 203 & 0.9158 & 324.03 \\
 & GPT-5.4-mini & 331/348 & 191 & 0.9511 & 328.23 \\
 & GPT-5.6-sol  & 252/428 & 163 & 0.5888 & 193.25 \\
\midrule
\makecell[l]{Foursquare NYC (Tag)}  & GPT-5-mini & 222/360 & 263 & 0.6167 & 569.68 \\
\makecell[l]{Foursquare NYC (Name)} & GPT-5-mini & 86/381  & 216 & 0.2257 & 607.78 \\
\bottomrule
\end{tabular}%
}

\vspace{-.3cm}
\end{table}
\begin{figure}[t]
    \centering
    \includegraphics[width=0.98\linewidth,trim=0.cm 0cm 0cm 0cm,clip]{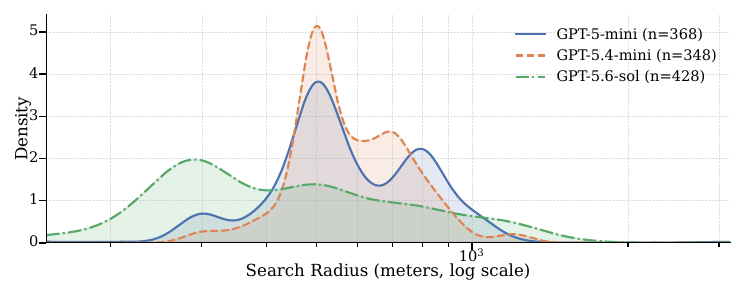}\vspace{-0.2cm}
    \includegraphics[width=0.99\linewidth,trim=0.cm 0cm 0cm 0cm,clip]{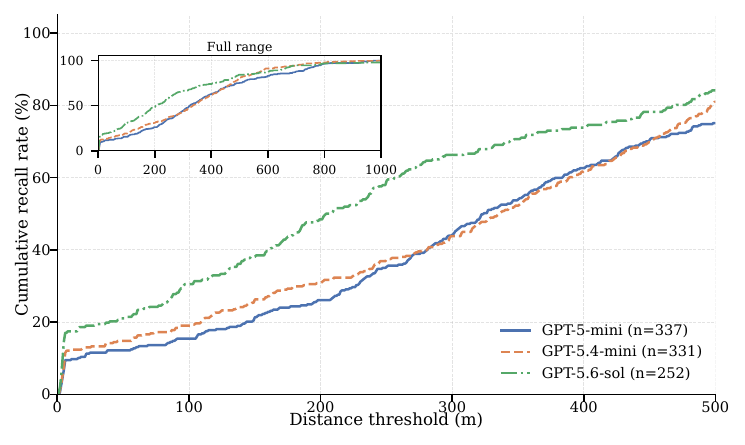}\vspace{-0.1cm}
    \captionsetup{width=1.\linewidth}\vspace{-0.4cm}
    \caption{Density distribution of model-suggested search radii (top) and cumulative recall rates for OSM tag search across varying distance thresholds (bottom).\vspace{-0.5cm}}
    \label{fig:osm_search_distance}
    \vspace{-0.1cm}
\end{figure}

\begin{table}[t]
\centering
\caption{Demographic Classification Performance across Agent Types and Models}
\vspace{-.3cm}
\label{tab:demographic_models_comparison}
\resizebox{\columnwidth}{!}{%
\begin{tabular}{@{}l l c c c@{}}
\toprule
\textbf{Agent Type ($N$)} & \textbf{Model} & \textbf{Predicted Precision} & \textbf{Predicted Recall} & \textbf{Predicted $F_1$ Score} \\
\midrule
\multirow{2}{*}{Homemaker (23)} 
 & GPT-5.4-mini & 0.5333 & 0.3478 & 0.4211 \\
 & GPT-5-mini & 0.6154 & 0.3478 & 0.4444 \\
 & GPT-5.6-sol & 0.8800 & 0.9565 & 0.9167 \\
\midrule
\multirow{2}{*}{Student (23)} 
 & GPT-5.4-mini & 1.0000 & 0.8261 & 0.9048 \\
 & GPT-5-mini & 1.0000 & 0.7826 & 0.8780 \\
 & GPT-5.6-sol & 1.0000 & 0.8261 & 0.9048 \\
\midrule
\multirow{2}{*}{Worker (54)} 
 & GPT-5.4-mini & 0.7385 & 0.8889 & 0.8067 \\
 & GPT-5-mini    & 0.7246 & 0.9259 & 0.8130 \\
 & GPT-5.6-sol & 0.9643 & 1.0000 & 0.9818 \\
\bottomrule
\end{tabular}
\vspace{-.5cm}
}
\vspace{-0.5cm}
\end{table}

\begin{table}[t]
    \caption{Token Usage Statistics for Selected Models}
    \vspace{-.3cm}
    \label{tab:token_usage}
    \centering
    \resizebox{\columnwidth}{!}{%
    \begin{tabular}{lccccccc}
        \toprule
        \textbf{Model} & 
        \textbf{Reqs.} & 
        \begin{tabular}{@{}c@{}}\textbf{In} \\ \textbf{Tokens}\end{tabular} & 
        \begin{tabular}{@{}c@{}}\textbf{Out} \\ \textbf{Tokens}\end{tabular} & 
        \begin{tabular}{@{}c@{}}\textbf{Cached} \\ \textbf{In Tokens}\end{tabular} & 
        \begin{tabular}{@{}c@{}}\textbf{Avg Tokens} \\ \textbf{In/Req}\end{tabular} & 
        \begin{tabular}{@{}c@{}}\textbf{Avg Tokens} \\ \textbf{Out/Req}\end{tabular} & 
        \begin{tabular}{@{}c@{}}\textbf{Avg Cost/} \\ \textbf{Req(\$)}\end{tabular} \\
        \midrule
        GPT-5-mini     & 100 & 1,363,012  & 298,596   & 0         & 13,630 & 2,986  & 0.003 \\
        GPT-5.4-mini   & 100 & 1,090,637  & 610,624   & 0         & 10,906 & 6,106  & 0.042 \\
        GPT-5.6-sol    & 100 & 7,571,825  & 3,250,699 & 2,732,446 & 75,718 & 32,507 & 0.416 \\
        \bottomrule
    \end{tabular}%
    }
    \vspace{-.5cm}
\end{table}

\begin{figure}[t]
    \centering
    \includegraphics[width=0.98\linewidth,trim=0cm 0cm 0cm 0cm,clip]{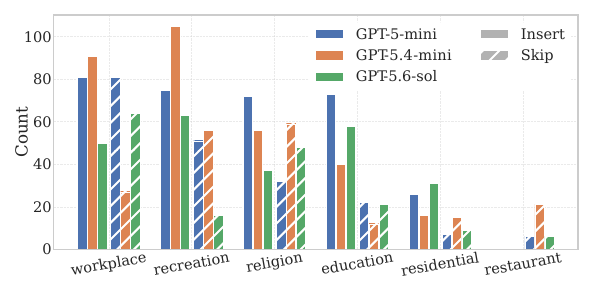}\vspace{-0.1cm}
    \captionsetup{width=1.\linewidth}\vspace{-0.2cm}
    \caption{Inserted \& skipped POIs count by category.\vspace{-0.1cm}}
    \label{fig:combined_insert_skip_by_category}
    \vspace{-0.5cm}
\end{figure}

\begin{figure*}[t]
    \centering
    \includegraphics[width=.99\linewidth,trim=0.0cm 0cm 0cm 0.0cm,clip]{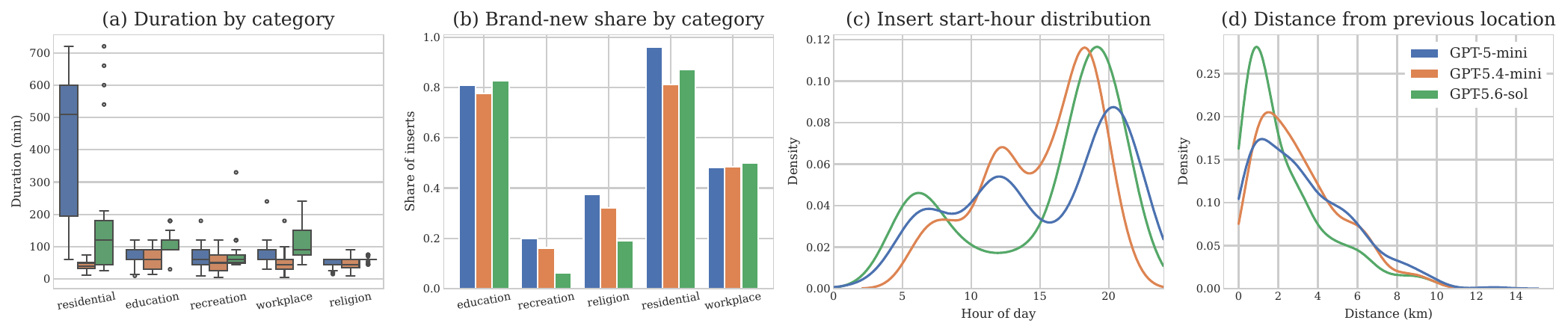}
    \vspace{-0.45cm}
    \captionsetup{width=1.\linewidth}
    \caption{Trajectory insertion characteristics across models. (a) Activity duration by category. (b) Proportion of brand-new locations per category. (c) Start-hour distribution. (d) Distance from previous location.\vspace{-0.45cm}}
    \label{fig:insert_eval}
    \vspace{-0.0cm}
\end{figure*}

\begin{figure}[t]
    \centering
    \includegraphics[width=0.98\linewidth,trim=0.25cm 0.2cm 0.2cm 0.2cm,clip]{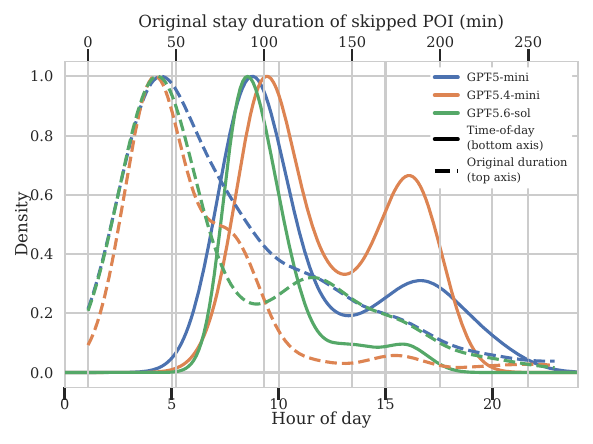}\vspace{-0.1cm}
    \captionsetup{width=1.\linewidth}\vspace{-0.2cm}
    \caption{Temporal and spatial characteristics of skipped visits. Solid lines (bottom axis): skip hour of day; dashed lines (top axis): stay duration.\vspace{-0.5cm}}
    \label{fig:skip_time_and_detour_combined}
    \vspace{-0.2cm}
\end{figure}


\begin{table*}[t]
    \small
    \centering
    \caption{Evaluation of Generated Abnormal Trajectories Across Perturbation Variants.\vspace{-0.2cm}}
    \begin{adjustbox}{width=1\textwidth,center}
        \begin{tabular}{@{}lcccccccc@{}}
            \toprule
            \multirow{3}{*}{\makecell{Trajectory \\ Perturbation Variant}} & \multirow{3}{*}{\makecell{Model}} & \multicolumn{7}{c}{{Evaluation Metrics}} \\
            \cmidrule(l){3-9}
            & & {\makecell{Kinematic \\ Violation Rate }}
            & {\makecell{Dwell-Time \\ Shift (s) }}
            & {\makecell{Spatial \\ Novelty Ratio }}
            & {\makecell{Normalized FastDTW \\ Deviation Cost (m) }}
            & {\makecell{Centroid Shift \\ Distance (m) }}
            & {\makecell{Transition \\ Surprisal ($\Delta \mathcal{L}$) }}
            & {\makecell{Temporal \\ Retention Rate }} \\
            \midrule
            Insert & \multirow{6}{*}{GPT-5 mini} & 0.0053 & -1.7082 & 0.0274 & 4.7207 & 6.7733 & 0.1172 & 1.0000 \\
            Skip &  & 0.0050 & 1.1762 & 0.0257 & 8.9523 & 15.4753 & 0.0268 & 1.0000 \\
            Insert w/Noise &  & 0.4260 & -5929.4957 & 0.2093 & 34.0098 & 6.8082 & - & 1.0000 \\
            Skip w/Noise &  & 0.4160 & -5931.2831 & 0.2079 & 43.1483 & 15.5058 & - & 1.0000 \\
            Insert+Skip &  & 0.0052 & -0.4581 & 0.0275 & 12.1071 & 14.9140 & 0.1432 & 1.0000 \\
            Insert+Skip w/Noise &  & 0.4258 & -5933.9265 & 0.2095 & 48.3594 & 14.9200 & - & 1.0000 \\
            \midrule
            Insert & \multirow{6}{*}{GPT-5.4 mini} & 0.0053 & -1.4878 & 0.0265 & 2.6384 & 4.1635 & 0.1083 & 1.0000 \\
            Skip &  & 0.0051 & -1.3924 & 0.0258 & 4.9522 & 6.9974 & 0.0106 & 1.0000 \\
            Insert w/Noise &  & 0.4262 & -5927.1635 & 0.2087 & 31.8521 & 4.1566 & - & 1.0000 \\
            Skip w/Noise &  & 0.4262 & -5929.2837 & 0.2086 & 35.8864 & 7.0353 & - & 1.0000 \\
            Insert+Skip &  & 0.0053 & -2.9796 & 0.0268 & 7.5501 & 7.8166 & 0.1192 & 1.0000 \\
            Insert+Skip w/Noise &  & 0.4262 & -5929.2831 & 0.2094 & 39.3105 & 7.8131 & - & 1.0000 \\
            \midrule 
            Insert & \multirow{6}{*}{GPT-5.6 sol} & 0.0052 & -0.9608 & 0.0270 & 2.5747 & 3.8251 & 0.1097 & 1.0000 \\
            Skip &  & 0.0051 & 0.8044 & 0.0254 & 7.3247 & 14.6472 & 0.0448 & 1.0000 \\
            Insert w/Noise &  & 0.4261 & -5928.4376 & 0.2094 & 31.7114 & 3.8554 & - & 1.0000 \\
            Skip w/Noise &  & 0.4260 & -5929.2651 & 0.2106 & 43.1613 & 14.6436 & - & 1.0000 \\
            Insert+Skip &  & 0.0052 & -0.1103 & 0.0268 & 10.9898 & 13.8334 & 0.1536 & 1.0000 \\
            Insert+Skip w/Noise &  & 0.4261 & -5932.3371 & 0.2094 & 46.3278 & 13.7760 & - & 1.0000 \\
            \midrule 
            Linear Interpolation & Baseline   & 0.0003 & 5714929.7163 & 0.0786 & 14.1236 & 17.0865 & - & 0.9663 \\
            Co-location Swap & Baseline   & 0.0064 & -0.0001 & 0.0603 & 81.7863 & 74.1249 & 0.1767 & 1.0000 \\
            \bottomrule
        \end{tabular}
    \end{adjustbox}
    \label{table:trajectory_eval}\vspace{-0.5cm}
\end{table*}


\subsubsection{Perturbation Variants}
To benchmark the proposed anomaly synthesis framework on the SF-Life dataset, we evaluate three LLMs: GPT-5-mini, GPT-5.4-mini, and GPT-5.6-sol Pro (abbreviated as GPT-5.6-sol). We use these models to synthesize anomalies across six distinct perturbation variants: \textbf{Insert-only}, \textbf{Skip-only}, \textbf{Insert with noise}, \textbf{Skip with noise}, \textbf{Combined Insert and Skip}, and \textbf{Combined Insert and Skip with noise}. At the API level, all models are constrained to a maximum length of 100,000 tokens. For GPT-5.6-sol, the reasoning effort is set to maximum.

Furthermore, we benchmark our framework's performance against two baseline perturbation methods:
\begin{enumerate}[leftmargin=*]
\item Linear Interpolation: This approach constructs synthetic trajectories from discrete check-in records. It first leverages check-in records to calculate the temporal transit durations between consecutive POI. A spatial trajectory is then generated by applying linear interpolation between these POI.
\item Co-location Swap: In this method, we swap the subsequent trajectory segments of two agents starting on the 30th day of the dataset once the algorithm identifies pairs of agents exhibiting spatial-temporal co-location.
\end{enumerate}
We restrict our comparison to these two approaches because the majority of existing generation methods are either closed-source or demand highly specific contextual data to model human intent. 
For example, frameworks that alter trajectories by applying psychological paradigms like Maslow's hierarchy of needs cannot be applied here, as the SF-Life dataset lacks the ultra-fine-grained semantic features required for such intent-based modifications. 
Conversely, because the Foursquare NYC dataset consists of discrete check-in records lacking raw, high-frequency trajectory data, we restrict its anomaly generation exclusively to the combined Insert + Skip variant.

\subsubsection{Evaluation Metrics}

To validate the quality of the synthesized dataset regarding kinematic plausibility and spatial validity, in addition to standardized statistical metrics and visualization methods, we utilize seven quantitative evaluation metrics to evaluate the generated trajectories.
As these metrics requires continuous high-frequency coordinate sequence input, we restrict their application exclusively to the SF-Life dataset.

\textbf{Kinematic Violation Rate (KVR):} evaluates the physical plausibility of the generated trajectory $T$ by analyzing the instantaneous velocity between consecutive points. It measures the proportion of spatial-temporal transitions that exceed the agent's maximum dynamic speed threshold, $v_{\text{max}}$, relative to the original trajectory. 

\textbf{Dwell-Time Shift (DTS):}
This metric evaluates the temporal stationarity of the modeled agent by quantifying the discrepancy in total stopping duration between the generated and original trajectories. In our case, an agent is considered dwelling if its instantaneous velocity $v_i$ falls below a dynamic dwell-speed threshold, $v_{\text{dwell}}$. 
This metric quantifies whether the synthetic trajectory includes excessive stationary periods compared to the original.

\textbf{Spatial Novelty Ratio (SNR):} This metric evaluates the efficacy of spatial anomaly injection by quantifying the proportion of newly explored geographical space.
We discretize the agent's routine mobility footprint into a 100x100 spatial heatmap, defining $\mathcal{G}_{\hat{T}}$ as the set of all populated grid cells derived from the ground-truth sequence $\hat{T}$.
The SNR computes the fraction of trajectories in the generated trajectories $T$ that fall either entirely outside the original bounding box or within historically unvisited grid cells.

\textbf{Normalized FastDTW (NDTW)~\cite{salvador2007toward}:}
This metric quantifies the absolute spatial divergence between the generated trajectory and the ground-truth sequence.
We explicitly exclude point-to-point metrics, such as Mean Distance Error, because our perturbation methodology introduces abnormal scheduling. 

\textbf{Centroid Shift Distance (CSD):}
This metric evaluates the macro-level geographic stability of the synthetic trajectory. 
A minimal Centroid Shift indicates that the injected anomaly functions as a realistic, localized detour. Conversely, a substantial shift signifies a long-distance spatial displacement, rendering the manipulated trajectory easily identifiable by standard anomaly detection models.

\textbf{Transition Surprisal ($\Delta \mathcal{L}$):} $\Delta \mathcal{L}$ quantifies semantic behavioral anomalies by measuring the information-theoretic deviation of a synthetic trajectory from historical mobility patterns. It is calculated as the difference in average negative log-likelihood between the modified and ground-truth sequences, utilizing first-order Markov transition probabilities $P(c_t \mid c_{t-1})$. A significant positive $\Delta \mathcal{L}$ mathematically validates the injection of statistically rare transitions.

\textbf{Temporal Retention Rate (TRR):} quantifies structural preservation by calculating the ratio between the cardinality of the generated synthetic trajectory $T$ and the original ground-truth $\hat{T}$.
This metric assesses frameworks that incur data loss from sampling, temporal masking, or sequence skipping.


\begin{figure*}[t]
    \centering
    
    \subfloat[with Macro-Environment Noise.]{%
        \includegraphics[width=0.30\linewidth]{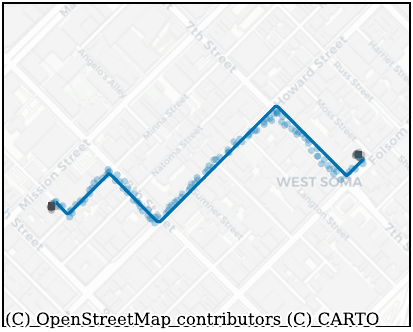}%
        \label{fig:noised_drift}}%
    \hspace{0.5em}%
    \subfloat[with Micro-Environment Noise.]{%
        \includegraphics[width=0.30\linewidth]{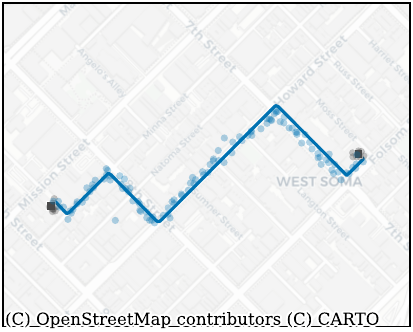}%
        \label{fig:rnoised_multipath}}%
    \vspace{-0.1cm}
    \subfloat[with Hardware Noise.]{%
        \includegraphics[width=0.30\linewidth]{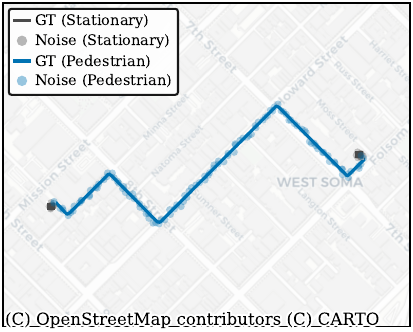}%
        \label{fig:noised_hardware}}%
    \hspace{0.5em}%
    \caption{Visualization of the three noise layers added to the raw data: (a) macro-environment noise, (b) micro-environment noise, and (c) hardware noise.}
    \label{fig:noise_layer}
    \vspace{-0.5cm}
\end{figure*}

\begin{figure*}[t]
    \centering
    \subfloat[Agent 71778 - Modality: Bicycle.]{%
        \includegraphics[width=0.30\linewidth]{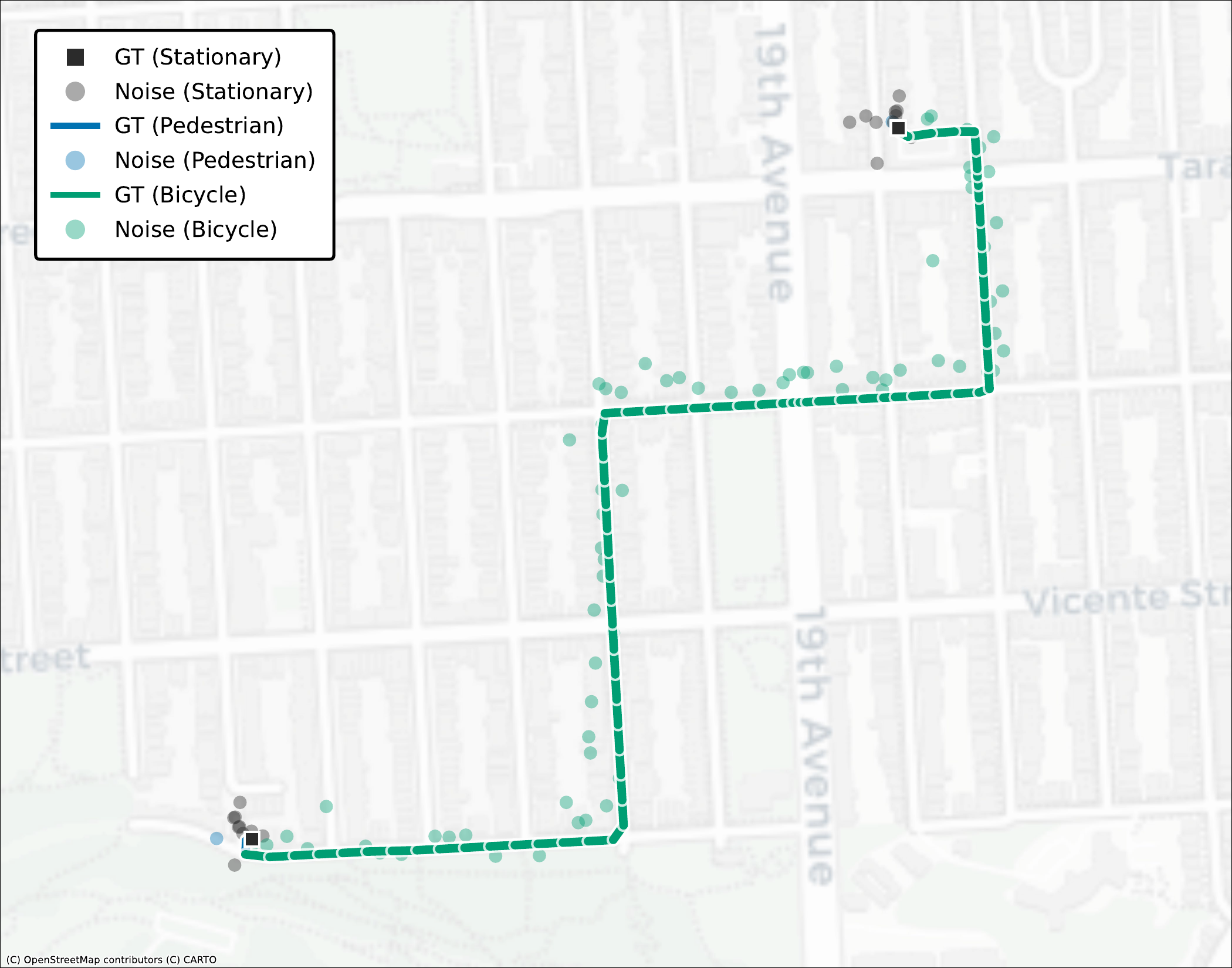}%
        \label{fig:noised_71778}}%
    \hspace{0.5em}%
    \subfloat[Agent 36720 - Modality: Pedestrian.]{%
        \includegraphics[width=0.30\linewidth]{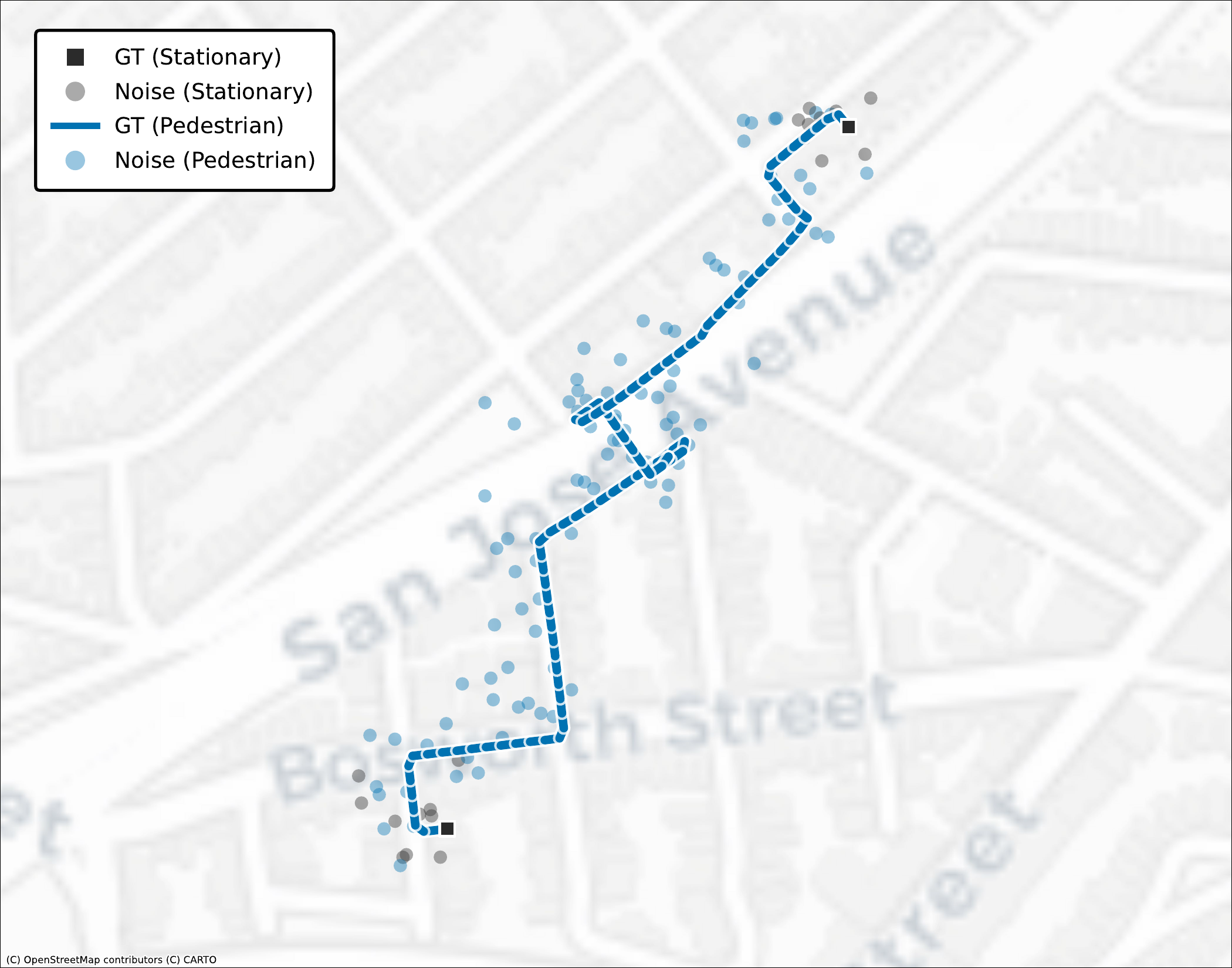}%
        \label{fig:noised_36720}}%
    \hspace{0.5em}%
    \subfloat[Agent 6556 - Modality: Auto.]{%
        \includegraphics[width=0.30\linewidth]{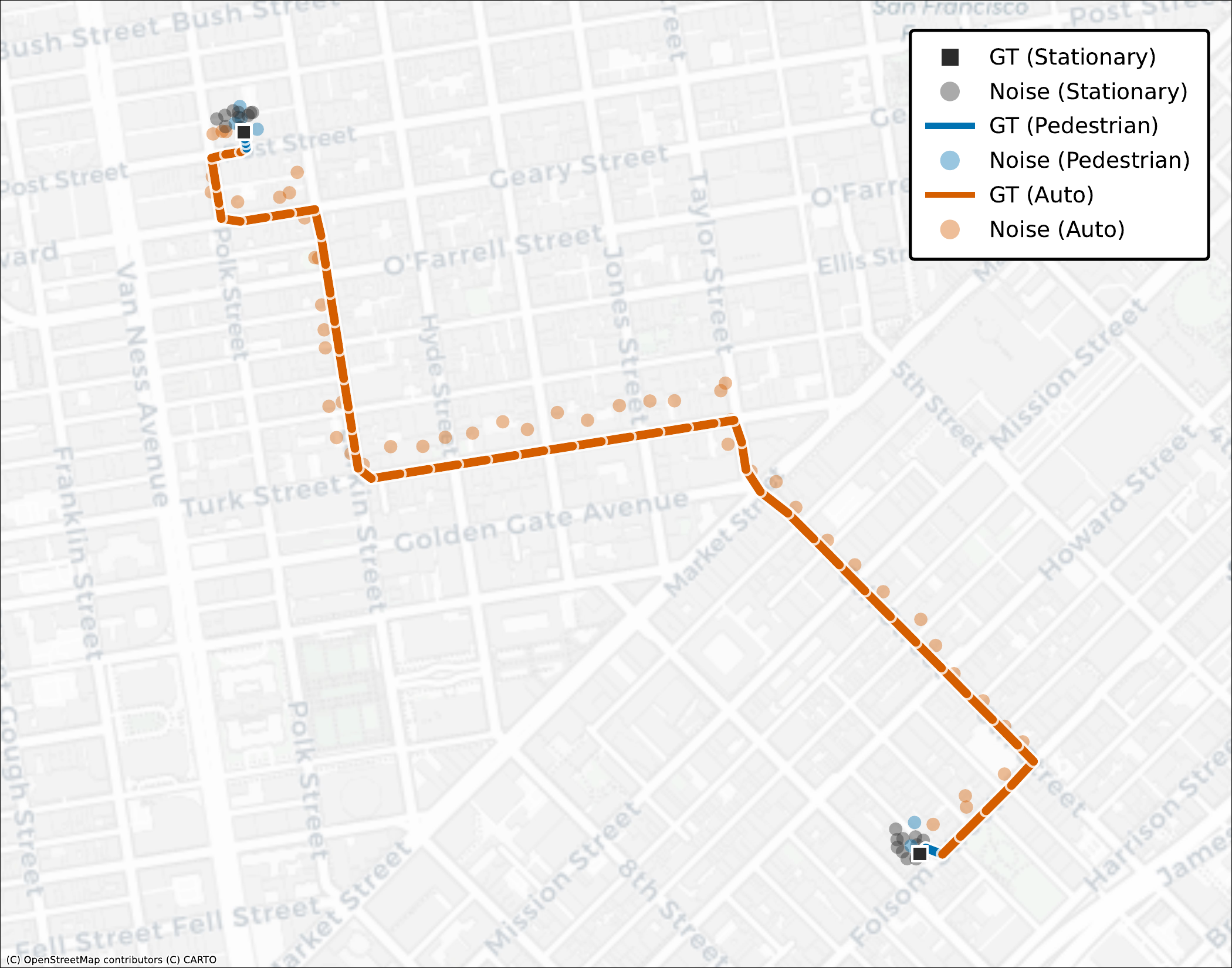}%
        \label{fig:rnoised_6556}}%
    \vspace{-0.1cm}
    \caption{Examples of modified agent trajectories with injected noise across various travel modes and distances.}
    \label{fig:noise}
    \vspace{-0.5cm}
\end{figure*}
\vspace{-0.3cm}
\begin{figure*}[t]
    \centering
    \vspace{-0.9cm}
    \subfloat{
        \centering
        \includegraphics[width=0.33\textwidth]{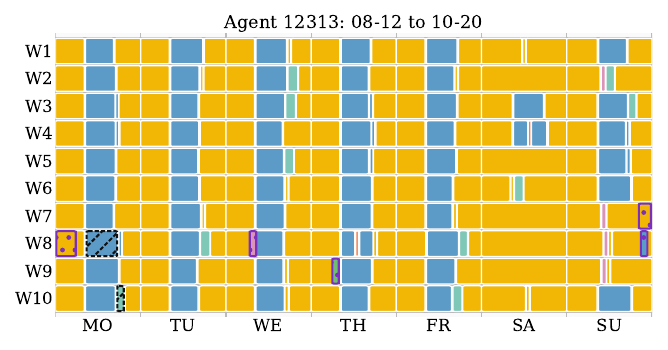}
    }%
    \subfloat{
        \centering
        \includegraphics[width=0.33\textwidth]{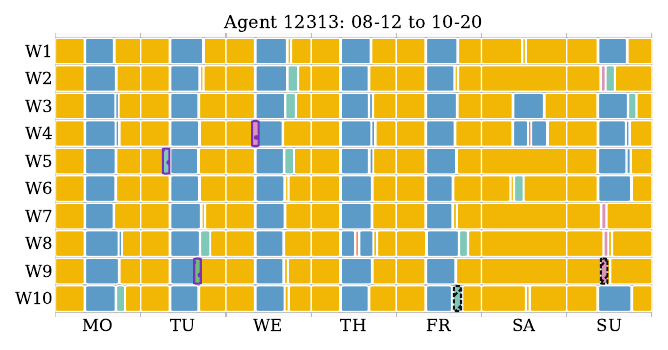}
    }%
    \subfloat{
        \centering
        \includegraphics[width=0.33\textwidth]{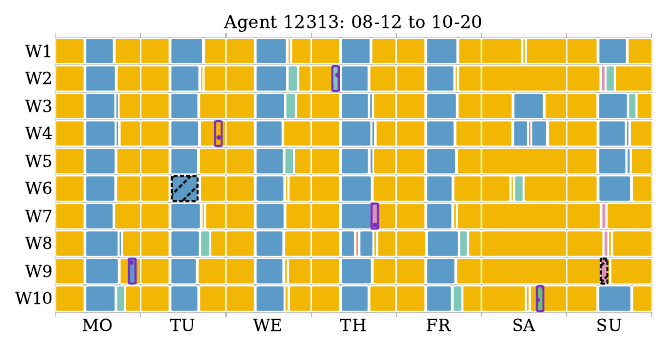}
    }
    
    \vspace{-0.5cm} 
    
    \subfloat{
        \centering
        \includegraphics[width=0.33\textwidth]{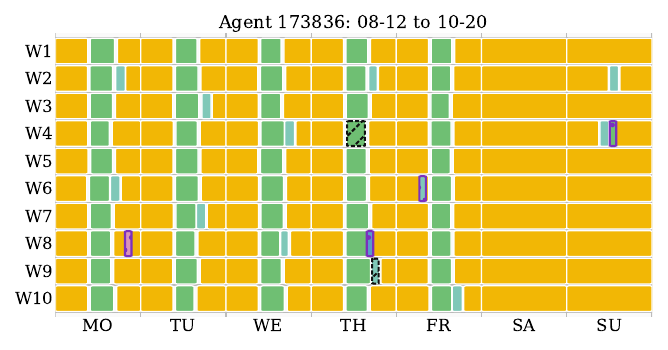}
    }%
    \subfloat{
        \centering
        \includegraphics[width=0.33\textwidth]{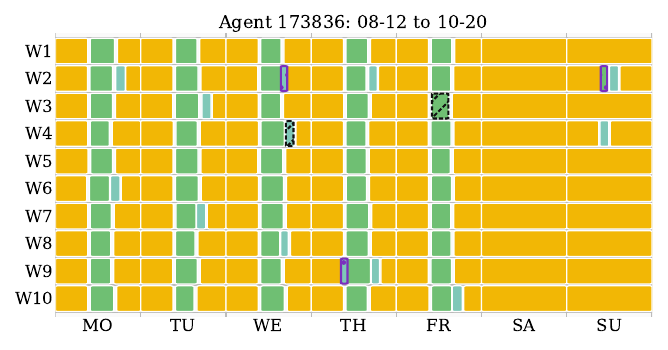}
    }%
    \subfloat{
        \centering
        \includegraphics[width=0.33\textwidth]{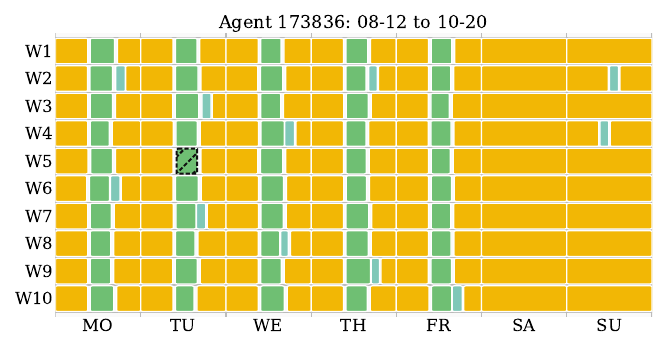}
    }
    
    \vspace{-0.5cm} 

    \makeatletter
    \setcounter{subfigure}{0}
    \makeatother
    \subfloat[GPT-5 mini]{
        \centering
        \includegraphics[width=0.33\textwidth]{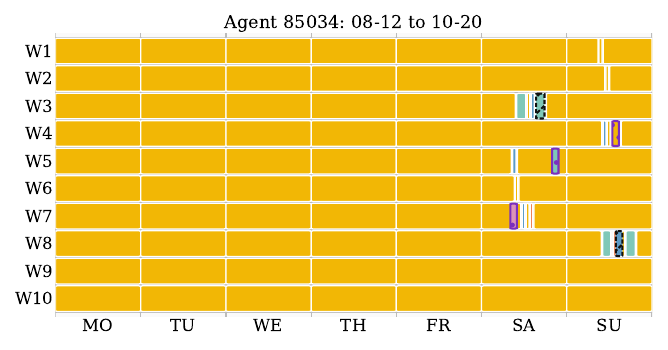}
        \vspace{-0.7cm}
    }%
    \subfloat[GPT-5.4 mini]{
        \centering
        \includegraphics[width=0.33\textwidth]{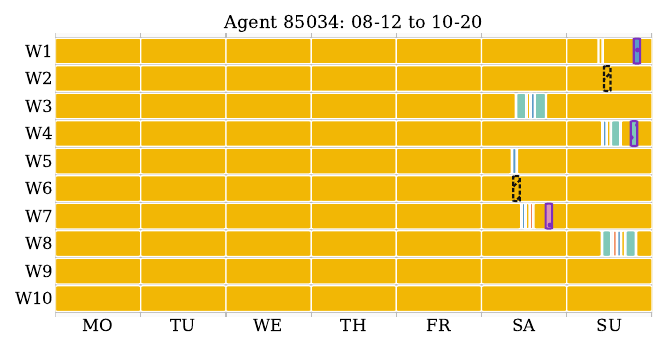}
        \vspace{-0.7cm}
    }%
    \subfloat[GPT-5.6 sol]{
        \centering
        \includegraphics[width=0.33\textwidth]{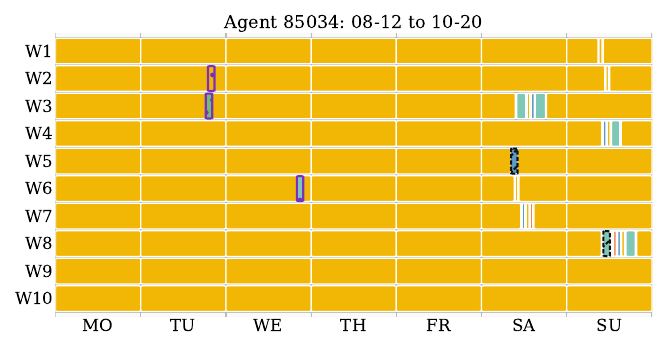}
        \vspace{-0.7cm}
    }
    \vspace{-0.39cm}
    \subfloat{
        \centering
        \includegraphics[width=\textwidth,trim=0cm 0.25cm 0cm 0.18cm,clip]{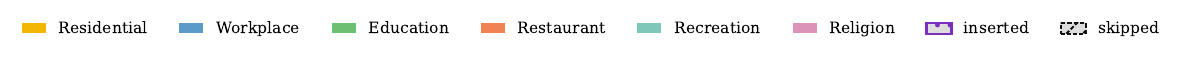}
    }
    \vspace{-0.4cm}
    \caption{Comparative analysis of schedule modifications across three distinct user profiles.}
    \label{fig:traj_cal_plot}\vspace{-0.7cm}
\end{figure*}

\subsection{Evaluation of LLM Perturbation Strategies}

Table~\ref{tab:LLM_Perturbation_statistics} provides a quantitative overview of the LLM perturbation and spatial mapping metrics across the three LLMs and datasets.

Within the SF-Life dataset, mapping efficacy and spatial constraints vary significantly across model architectures. Both mini models (GPT-5-mini and GPT-5.4-mini) match at median radii of 324.03 m and 328.23 m, respectively. These broader search spaces facilitate high spatial alignment, achieving insert resolve rates of 0.9158 and 0.9511. In contrast, GPT-5.6-sol, configured with maximum reasoning effort, enforces a substantially shorter median match radius of 193.25 m. As illustrated in Figure~\ref{fig:osm_search_distance}, its density distribution skews heavily toward these shorter distances. This tight focus yields over 80\% recall within 500 m; however, because such small bounding radii often contain no matching OSM POI, GPT-5.6-sol resolves only 252 of its 428 planned insertions (resolve rate 0.5888).
These discrepancies suggest that the model-suggested search radius functions as a proxy for internal predictive confidence rather than genuine spatial awareness. Advanced architectures, such as GPT-5.6-sol, generate tightly constrained search radii. Because LLMs lack accurate geographic awareness, this restricted search space frequently fails to encircle valid OSM nodes, leading to lower resolve rates. Conversely, the less confident mini models output substantially larger search radii, inadvertently compensating for the absence of structural spatial intelligence by capturing a greater number of valid nodes and yielding higher resolve rates.

The distribution of inserted and skipped POIs varies significantly across both POI categories and model architectures. As detailed in Figure~\ref{fig:combined_insert_skip_by_category}, the workplace and recreation categories attract the highest volume of insertions uniformly across all models, whereas categories such as religion and education exhibit a pronounced divergence in skip frequencies. Furthermore, evaluating the proportion of brand-new locations introduced per category (Figure~\ref{fig:insert_eval}b) reveals that the models mainly inject novel venues within residential and educational contexts, while relying on historically familiar or recurrent spatial templates for recreational and religious activities.

Spatially, the displacement from preceding locations (Figure~\ref{fig:insert_eval}d) demonstrates that the majority of injected events occur within a constrained 0 to 2km window, a localization trend that is particularly noticeable in the GPT-5.6-sol architecture. Temporally, activities inserted by GPT-5.6 sol exhibit a sharper early-morning peak compared to those generated by the mini variants (Figure~\ref{fig:insert_eval}c). Conversely, the analysis of skipped visits (Figure~\ref{fig:skip_time_and_detour_combined}) indicates that the skip perturbation primarily targets the mid-morning activities, peaking between hours 8 and 10. The original stay durations for these skipped visits are tightly concentrated around 40 minutes, suggesting that the models preferentially excise brief, transitional stops rather than extended core activities.

\subsection{OSM Search Strategy Analysis}
To mitigate LLM spatial hallucinations, we benchmark two spatial resolution strategies on the Foursquare NYC dataset using OSM Tag Search, which queries OpenStreetMap using category name, and OSM Name Search, which directly queries specific entity titles generated by the LLM. 
As shown in Table~\ref{tab:LLM_Perturbation_statistics}, OSM Tag Search yields an insert resolve rate of 61.67\%, significantly outperforming OSM Name Search at 22.57\%.
Furthermore, OSM Tag Search finds its matched POIs at a slightly smaller median radius than OSM Name Search (569.68 m vs. 607.78 m).
These findings confirm that LLMs lack real-world spatial awareness and frequently name non-existent venues. In contrast, our OSM tagging approach resolves nearly three times as many insertions, substantially mitigating the occurrence of hallucinations.

\vspace{-0.2cm}
\subsection{LLM Context Awareness Experiment}
To validate the LLM's capability to deduce agent occupations and underlying behavioral patterns from mobility trajectories, Table~\ref{tab:demographic_models_comparison} shows its demographic classification performance on the SF-Life dataset. GPT-5.6 sol matches or outperforms the mini models in every category.
Students exhibit the most identifiable patterns, with all models achieving perfect precision (1.0000) and F1 scores above 0.87. Homemakers prove most challenging for mini models (recall 0.3478, F1 $<$ 0.45); however, the sol model resolves this deficiency, reaching an F1 score of 0.9167. For Workers, the sol model achieves a near-perfect classification ($F_1 = 0.9818$), surpassing the mini models' baseline ($F_1 \approx 0.81$).
\vspace{-0.2cm}
\subsection{Noise Module Evaluation}
As absolute ground truth does not exist for the synthesized spatial noise, we rely on qualitative visual inspections to validate the noise module.
Figure~\ref{fig:noise_layer} illustrates the noise module described in Section~\ref{sec:noise} progressively compounding impact of macro-environment, micro-environment, and hardware noise, which jointly introduce significant spatial distortion to the trajectory. 
Figure~\ref{fig:noise_layer}c isolates the baseline hardware noise, where positional deviations remain minimal and tightly clustered along the ground truth. 
Figure~\ref{fig:noise_layer}a introduces atmospheric delays derived from weather data~\cite{cds.adbb2d47,GNSS_IGSIONOTEC} and AR(1) drift to this baseline; the AR(1) process temporally correlates the error magnitude, ensuring that the dispersion of positional errors evolves smoothly with time-varying tropospheric and ionospheric conditions rather than uniform, uncorrelated scatter. 
Finally, Figure~\ref{fig:noise_layer}b demonstrates the severe trajectory degradation characteristic of dense urban environments through the injection of heavy-tailed micro-environmental noise, which results in extreme positional outliers.
Building upon this, Figure~\ref{fig:noise} validates the noise model across bicycle, pedestrian, and automobile modalities, demonstrating an inverse relationship between agent velocity and trajectory degradation.
Specifically, the low-velocity pedestrian modality (Figure~\ref{fig:noise}b) exhibits severe relative spatial variance because time-correlated and localized multipath errors accumulate densely over short distances. Conversely, the high-velocity automobile trajectory (Figure~\ref{fig:noise}c) maintains a smoother observed path.
As the plotted trajectories demonstrate, the proposed framework synthesizes realistic, spatially heterogeneous noise. Unlike spatially homogeneous noise, this design avoids uniform scatter and the detectable artifacts it produces.

\vspace{-0.3cm}
\subsection{Final CTI Trajectory Analysis}
\subsubsection{Quantitative Results}
Table \ref{table:trajectory_eval} presents the evaluation metrics across the three generative models (GPT-5 mini, GPT-5.4 mini, and GPT-5.6 sol). Un-noised variants (Insert, Skip, and Insert+Skip) maintain an exceptionally low KVR (0.0050–0.0053) across all models, matching the Co-location Swap baseline and preserving physical plausibility without teleportation artifacts. Conversely, noised variants exhibit elevated KVRs strictly due to coordinate jittering. 
Applying random spatial noise to discrete trajectory points artificially alters the distance traveled between consecutive timestamps, which generates synthetic spikes in the calculated instantaneous velocity and acceleration that exceed physical kinematic thresholds. 
All proposed variants achieve a TRR of 1.0000, ensuring lossless data collection in the synthesized dataset.

Regarding DTS, un-noised variants exhibit minimal shifting (-2.98 to 1.18 s), preserving the agent's total stationary duration. Noised variants induce severe DTS artifacts because metric calculations erroneously register coordinate fluctuations as continuous drift. For the SNR metric, un-noised variants yield localized, modest deviations of 0.026, while noise injection scales the value to 0.2106. 
This result shows that our perturbation rules introduce noticeable spatial deviations while keeping visits within realistic geographic regions.
Moreover, the positive transition surprisal (0.011–0.154) across all un-noised variants confirms the injected transitions represent true routine deviations rather than random noise.

Geometric similarity highlights subtle model variations. GPT-5.4 mini demonstrates the tightest spatial control, yielding the lowest NDTW (7.5501 m) and CSD (7.8166 m) under combined Insert+Skip operations. Across all models, un-noised variants bound global spatial displacement strictly under 16 m, confirming the LLMs generate realistic, non-overlapping, and localized detours.

In contrast, the baselines produce severe structural artifacts. Co-location Swap incurs massive NDTW (81.7863 m) and CSD (74.1249 m) errors, representing highly unnatural global migration. Linear Interpolation artificially constrains deviations by targeting only a minor fraction of transit segments. Consequently, both baseline methods introduce global structural distortions easily detectable by standard anomaly detection models.

\subsubsection{Qualitative Results}
We conducted a qualitative visual inspection on several generated anomalies to verify structural and physical plausibility.
Visual evidence from Figure \ref{fig:samples} demonstrates the framework's capability to seamlessly execute behavioral modifications. 
In the insertion scenarios, the synthetic trajectory successfully diverges from the original POI $A$ to incorporate the newly inserted POI $B$ without exhibiting unnatural discontinuities or sudden spatial jumps (``teleportations'') while adhering to physical constraints of the road network. 
Similarly, for skip modifications, the generated trajectory excludes the skipped trip to and from Location $B$ and creates a new trajectory directly from $A$ to $C$ that, qualitatively, appears difficult to discriminate from any trajectory originally found in the dataset. 
Meanwhile, we visualize the spatial-temporal trajectories of three representative agents via calendar plots in Figure~\ref{fig:traj_cal_plot}. 
In this figure, we present three agent samples: worker (Agent 12313), student (Agent 173836), and homemaker (Agent 85034).
All models introduce localized perturbations while preserving the agents' core weekly structures. Workers retain their Monday–Friday commutes alongside minor schedule adjustments. Students maintain their core weekday educational blocks, with modifications confined to peripheral activities. Similarly, homemakers exhibit sparse, low-frequency perturbations (e.g., occasional weekend errands) that align with their limited mobility profiles, ensuring all behavioral changes remain contextually plausible.
These visualizations confirm that the injected anomalies do not consist of stochastic, unconstrained insertions or deletions. Rather, the fundamental mobility semantics of the agents are preserved without introducing unrealistic behavioral discontinuities.

\begin{table}[t]
    \centering
    \caption{Downstream task performance of anomaly detection models across different perturbation models}
    \label{tab:perturbation_metrics}
    \resizebox{\columnwidth}{!}{%
    \begin{tabular}{clccc}
        \toprule
\multirow{2}{*}{\makecell{Perturbation \\ Model}} & 
        \multirow{2}{*}{\makecell{Abnormal \\ Detection Model}} & 
        \multicolumn{3}{c}{Metrics} \\
        \cmidrule(lr){3-5} 
        & & AP & AUPRC & AUROC \\
        \midrule
        
        \multirow{4}{*}{GPT-5 mini} & GM-VSAE & 0.6142 & 0.6138 & 0.9427 \\
                                        & NCF    & 0.4271 & 0.4289 & 0.8932 \\
                                        & USTAD  & 0.7483 & 0.7479 & 0.9714 \\
                                        & ICAD   & 0.5916 & 0.5904 & 0.9365 \\
        \midrule
        \multirow{4}{*}{GPT-5.4 mini} & GM-VSAE & 0.5873 & 0.5869 & 0.9284 \\
                                        & NCF    & 0.3856 & 0.3962 & 0.8651 \\
                                        & USTAD  & 0.7312 & 0.7280 & 0.9663 \\
                                        & ICAD   & 0.5781 & 0.5709 & 0.9155 \\
        \midrule                                
        \multirow{4}{*}{GPT-5.6 sol} & GM-VSAE & 0.5904 & 0.5919 & 0.9293 \\
                                        & NCF    & 0.3913 & 0.3987 & 0.8710 \\
                                        & USTAD  & 0.7185 & 0.7184 & 0.9592 \\
                                        & ICAD   & 0.5817 & 0.5797 & 0.9208 \\
        \bottomrule
    \end{tabular}%
    }\vspace{-0.5cm}
\end{table}
\vspace{-.2cm}

\subsection{Downstream Anomaly Detection Task Evaluation} \label{sec:downstream_performance} 

To evaluate the practical utility of the generated mobility benchmark, we deploy four anomaly detection methods: USTAD~\cite{wen2025uncertaintyawarehumanmobilitymodeling}, GM-VSAE~\cite{liu2020online}, ICAD~\cite{azarijoo2025icad}, and NCF~\cite{10.1145/3681765.3698463} across three noised datasets, each combining 10,000 unmodified SF-Life agents as normal agents with 100 anomalous agents perturbed by our framework. As detailed in Table~\ref{tab:perturbation_metrics}, detection performance is quantified via AP, AUPRC, and AUROC. 
Compared to the anomalies in NUMOSIM (GM-VSAE AUROC 0.507~\cite{stanford2024numosimsyntheticmobilitydataset}), detectors achieves an AUROC of around 0.9 and AP scores ranges from 0.39 (NCF) to 0.75 (USTAD) on our data, indicating the generated anomalies are more learnable yet fundamentally non-trivial.
Notably, downstream performance shows minimal variance between GPT-5.6-sol and the mini models. As evidenced by the low resolve rate of name-based POI search in Table~\ref{tab:LLM_Perturbation_statistics} (22.57\%) and the weekly routines shown in Fig~\ref{fig:traj_cal_plot}, current LLMs demonstrate strong temporal reasoning but lack precise real-world spatial awareness. Therefore, deploying smaller models introduces no significant performance degradation while still generating a balanced, learnable benchmark.

\vspace{-.1cm}

\subsection{Experiment Cost}
Table \ref{tab:token_usage} details token consumption and costs for a batch of 100 requests (one per agent). The mini models demonstrate high efficiency, costing \$0.003 (GPT-5-mini) and \$0.042 (GPT-5.4-mini) per request. Conversely, GPT-5.6-sol processes approximately ten times the token volume. Despite leveraging 2.7 million cached input tokens, this overhead drives its average cost to \$0.416 per request.
\vspace{-0.2cm}
\section{Conclusions and Future Work}
\label{sec:conclusion}
In this paper, we propose a novel end-to-end generative framework to address the critical scarcity of annotated ground-truth data in human trajectory anomaly research. This framework provides a scalable mechanism to generate high-fidelity annotated mobility datasets for the robust evaluation of downstream tasks.
By utilizing an LLM conditioned on each agent's mobility patterns and demographics, the framework systematically applies contextually grounded and physically feasible deviations into normative mobility trajectories via Insert and Skip operations.
To mitigate the LLM spatial hallucinations, we designed a robust three-stage translation pipeline. This pipeline integrates OSM-based POI resolution, demographics-aware travel-mode prediction, and map-constrained routing to convert discrete semantic anomaly events into continuous 0.2Hz GPS coordinate traces.
Finally, empirical downstream anomaly detection results demonstrate that our framework avoids the extremes of generating trivially simple or impractically difficult anomalies, thereby yielding a challenging yet learnable environment to reliably evaluate state-of-the-art architectures.
In future work, we will focus on modeling multi-agent collective anomalies, developing adversarial generation loops based on downstream detector feedback, and calibrating spatial noise against real-world GNSS traces.




\ifCLASSOPTIONcaptionsoff
  \newpage
\fi
\section*{Acknowledgment}

During the conduct of this research and the preparation of this manuscript, the authors utilized Large Language Models (LLMs), specifically OpenAI GPT series models, as a core methodological component of the proposed generative framework to plan synthetic mobility anomalies (\textit{Insert} and \textit{Skip} check-in perturbations conditioned on agent demographic personas) in Sections III and IV. Generative AI assistance was also used for grammar \& text refinement during manuscript preparation. The authors have thoroughly reviewed, verified, and audited all synthetic datasets, experimental outputs, and written text, and accept full responsibility for the accuracy, novelty, and scientific integrity of the work.



%
\bibliographystyle{IEEEtran}
\bibliography{refs/main}

%








\end{document}